\documentclass[journal]{IEEEtran}
\usepackage[utf8]{inputenc}

\usepackage{array}
\usepackage{ifpdf}
\usepackage{epsfig} % for postscript graphics files

\usepackage[cmex10]{amsmath}
\usepackage{amsfonts}
\usepackage{amsthm}
\usepackage{algorithm}
\usepackage{graphicx,float}
\usepackage{amsgen}
\usepackage{amssymb}
\usepackage{amsbsy}
\usepackage{mdwmath}
\usepackage{graphicx}
\usepackage{color}
\usepackage{colortbl}
\usepackage{hhline}
\usepackage{multirow}
\usepackage[caption=false,font=footnotesize]{subfig}
\usepackage{url}
\usepackage{fixltx2e}
\usepackage{float}
\usepackage{subfig}
\usepackage{cite}
\usepackage{balance}
\usepackage{todonotes}
\usepackage{booktabs}
\usepackage[english]{babel}
\newcommand{\beps}{\boldsymbol{\varepsilon}}
\newcommand{\eps}{{\varepsilon}}
\newcommand{\bell}{\boldsymbol\ell}
\newtheorem{prop}{Proposition}

\newtheorem{lemma}{Lemma}

\usepackage[noend]{algpseudocode}
\usepackage{hyperref}
\newcommand{\graphicalModel}{
\begin{figure}[t!]
\centering
\includegraphics[width=0.7\linewidth]{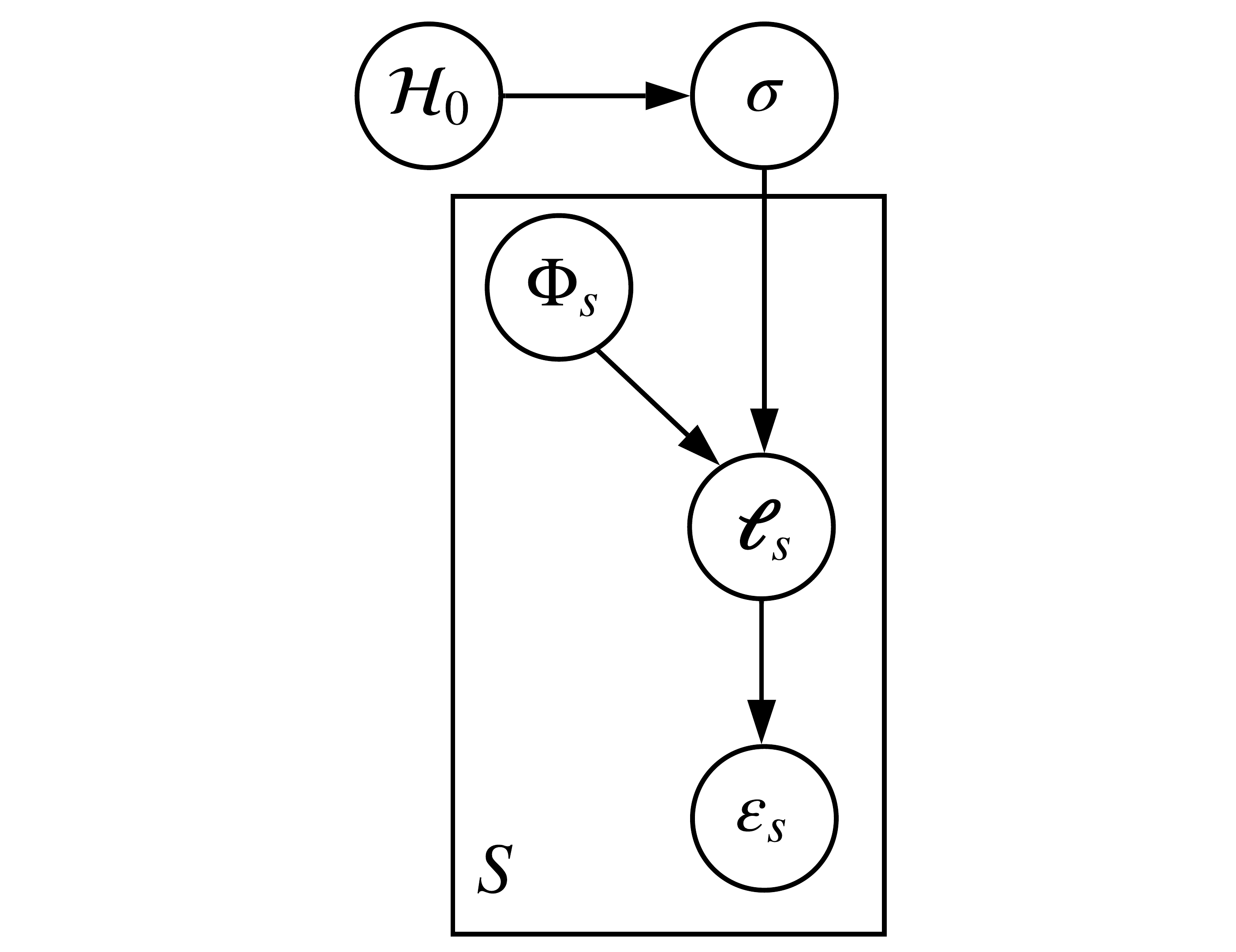}
\caption{{Proposed probabilistic graphical model representing the $S^\text{th}$
sequence for the RBI problem. $\mathcal{H}_0$ denotes prior information in the estimation. state $\sigma$ and the query set $\Phi_s$ generates the evidence $\beps_s$ through a label assignment $\bell_s$. $\bell_s$ defines the relationship between queries and the state in a probabilistic manner.}}
\label{fig:graphicalModel}
\end{figure}
}

\newcommand{\SimplexProbTrajectories}{
    \begin{figure*}[t!]
      \centering
      \subfloat[]{\label{fig:simplex_initialpoint}\includegraphics[width=0.25\textwidth, trim=0in 0in 0in 0.2in, clip =False]{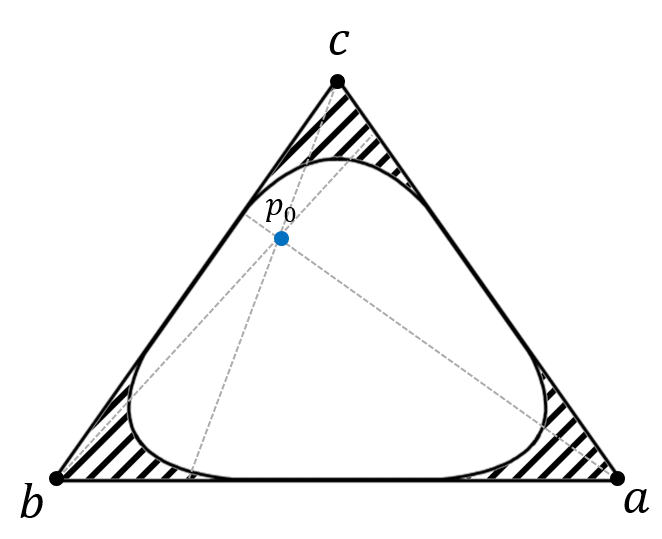}}
        \subfloat[]{\label{fig:simplex_proterior}\includegraphics[width=0.25\textwidth, trim=0in 0in 0in 0.2in, clip =False]{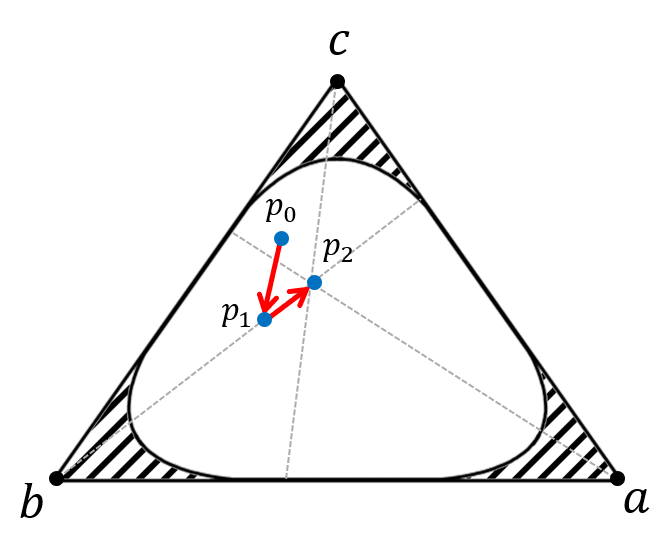}}
        \subfloat[]{\label{fig:simplex_probChanges}\includegraphics[width=0.25\textwidth, trim=0.1in 0in 0in 0.2in, clip =False]{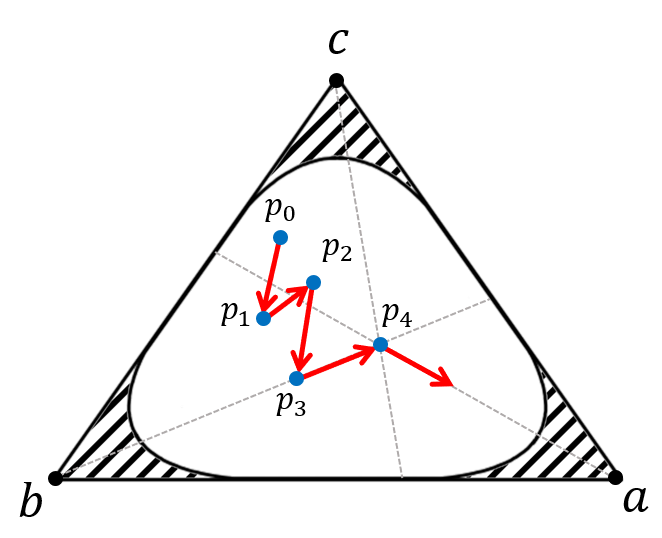}}
        \subfloat[]{\label{fig:simplex_stopping}\includegraphics[width=0.25\textwidth, trim=0in 0in 0in 0.2in, clip =False]{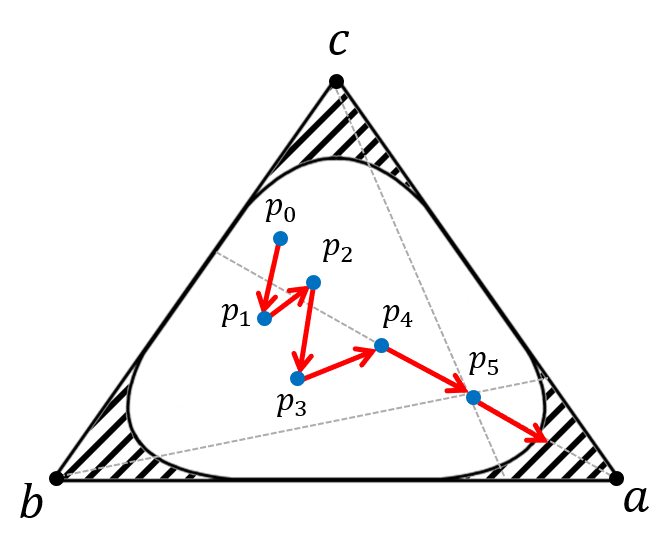}}
        \caption{{Illustration of posterior trajectories changes over sequences in RBI to reach the target vertex $\textbf{\textit{a}}$. (a) The dashed areas illustrate an example of a feasible region of the inference problem and the blue dot represents the initial probability over states. (b) Posterior changes over two sequences (queries). (c) After collecting sufficient evidence in four sequences, the system starts to move toward the target vertex. (d) Still not located in the feasible region of the target, however, based on the direction and speed of posterior changes, the system can foresee the location of the next move and terminate the process.}}
        \label{fig:simplex_trajectory}
\end{figure*}
}

\newcommand{\simulationAlpha}{
    \begin{figure*}[t!]
       \centering
      \includegraphics[width=0.9\linewidth]{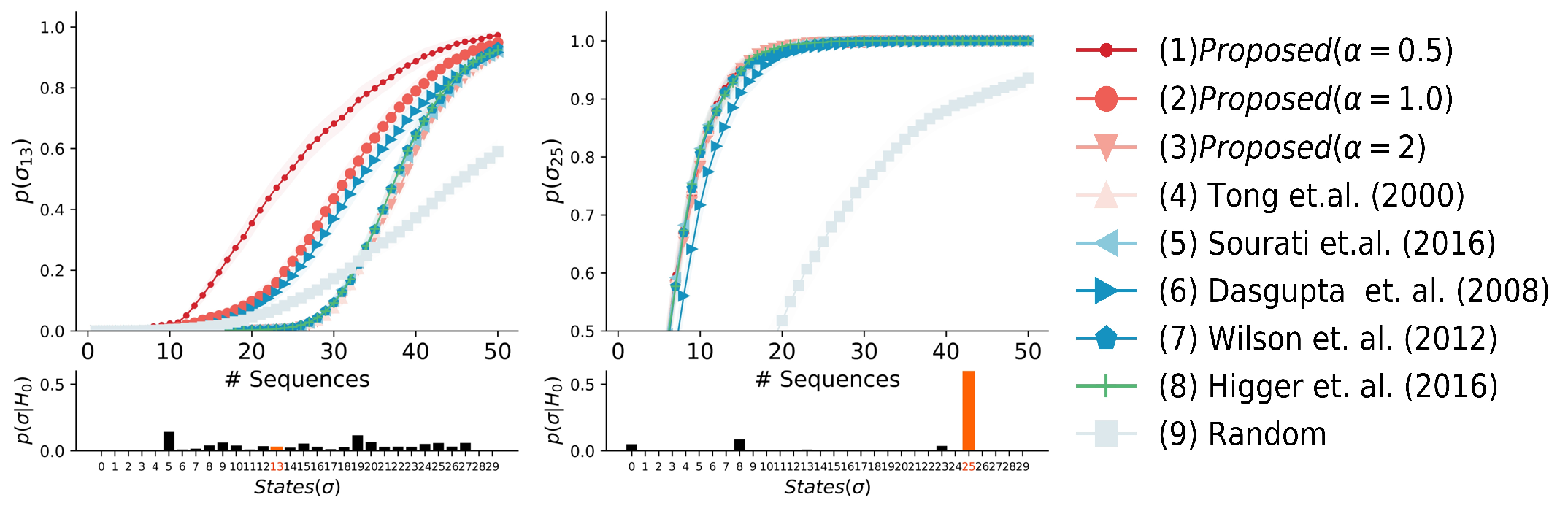}
       \caption{The impact of $\alpha$ on decision making. Rényi entropy based method for different $\alpha$ values is compared to \textit{random} and 5 other query selection methods in adversarial and supportive prior (from left to right) presence. Observe that, proposed methods increase the probability mass on the true estimate earlier and faster than other methods. Additionally, with increasing $\alpha$ values, estimation curves for the proposed method converge to \textit{N-best} methods curve.}
       \label{fig:simulation_diff_alpha}
    \end{figure*}
}

\newcommand{\foodRecommender}{
    \begin{figure}[t!]
      \centering
        \subfloat[]{\label{fig:food_supportive}\includegraphics[width=0.24\textwidth, trim=0in 0in 0in 0.2in, clip =False]{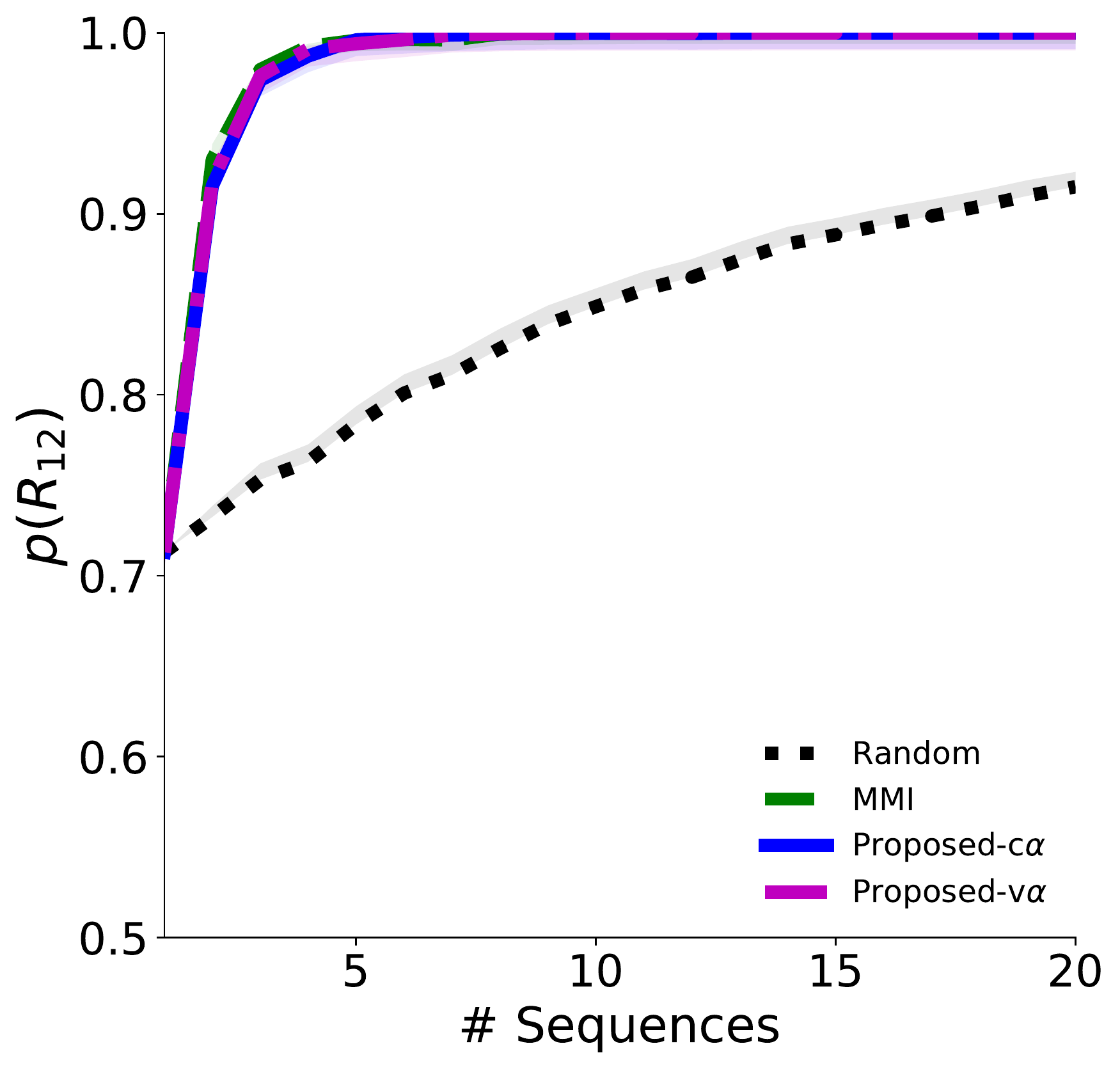}}
        \subfloat[]{\label{fig:food_adversarial}\includegraphics[width=0.24\textwidth, trim=0.1in 0in 0in 0.2in, clip =False]{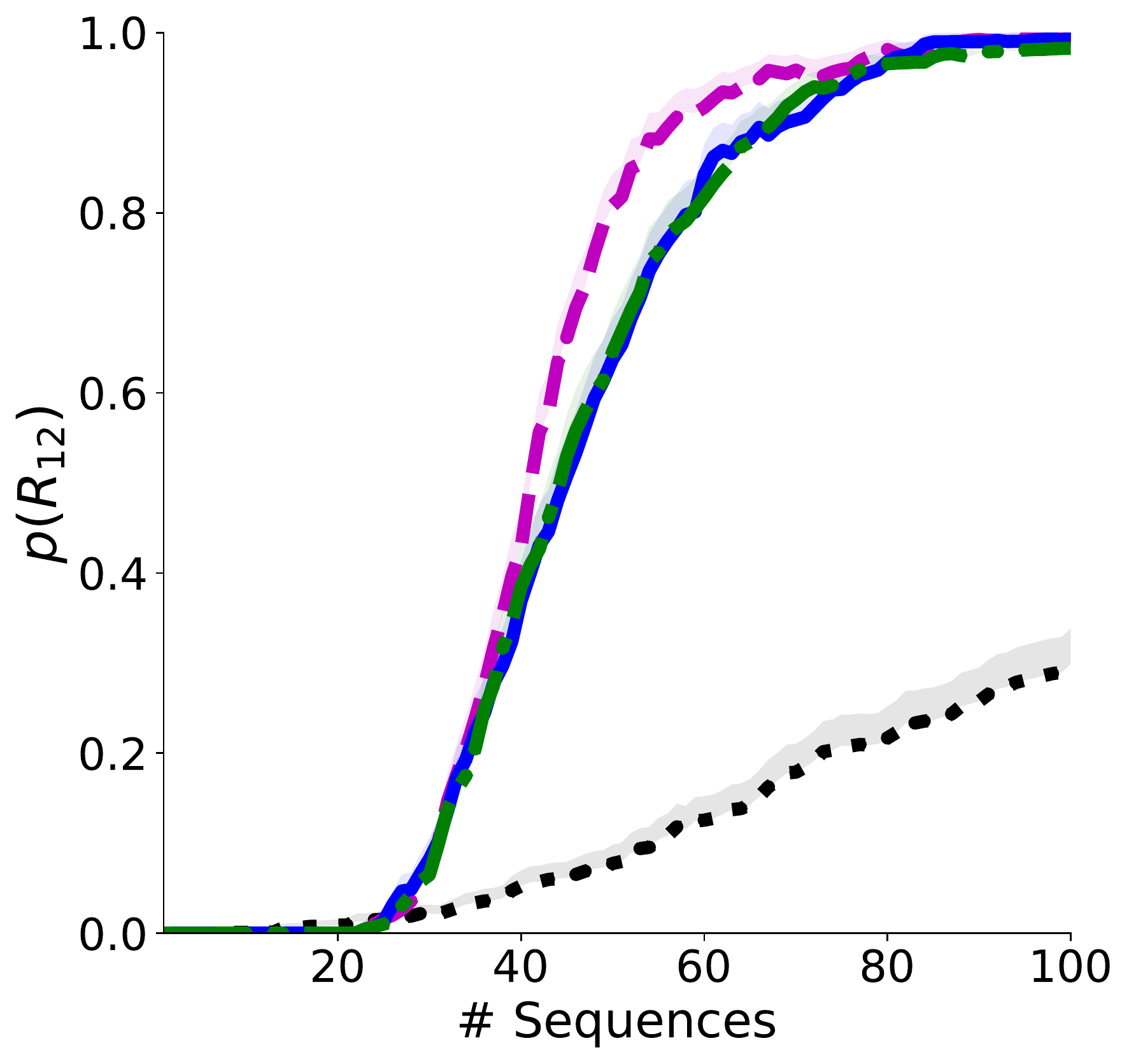}}
        \caption{{The intended restaurant probability changes in RBI for two users, (a) a conventional diner and (b) an adventurous diner.}}
        \label{fig:food_recommender}
\end{figure}
}

\newcommand{\BCItyping}{
    \begin{figure}[h!]
      \centering
        \subfloat[]{\label{fig:ITL}\includegraphics[width=0.25\textwidth, trim=0in 0in 0in 0.2in, clip =False]{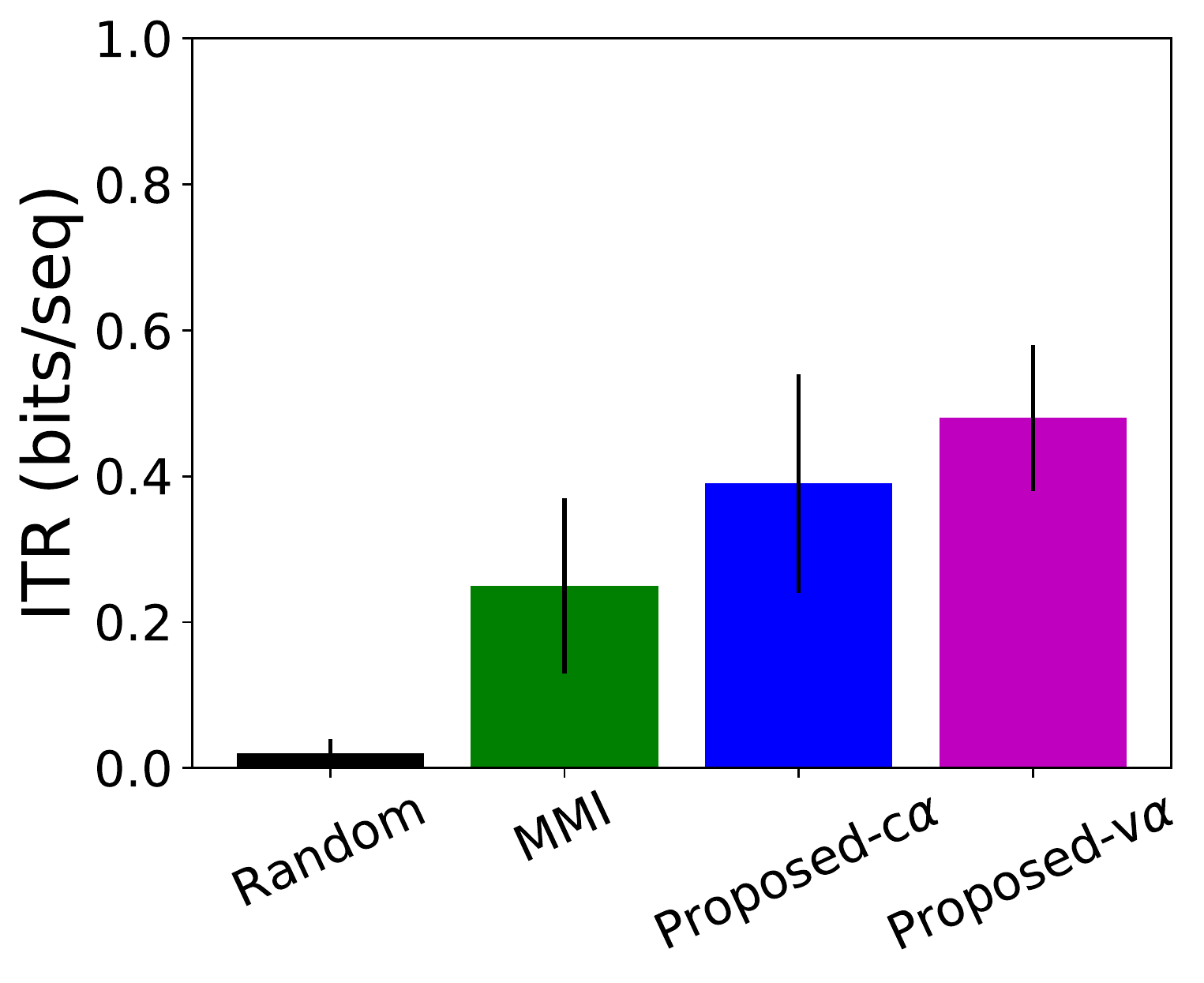}}
        \subfloat[]{\label{fig:ATL}\includegraphics[width=0.25\textwidth, trim=0.1in 0in 0in 0.2in, clip =False]{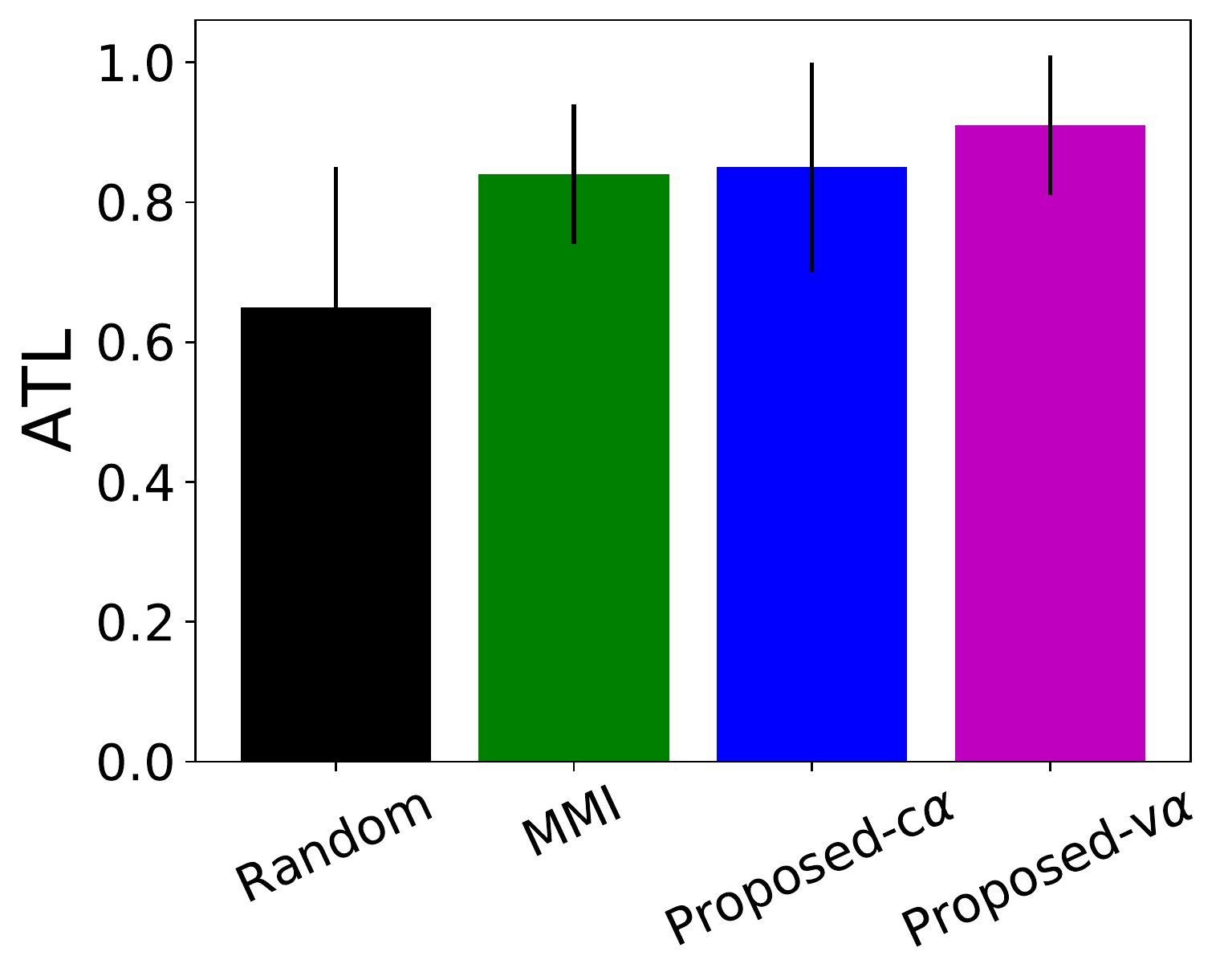}}
        \caption{{(a) Average of ITR and (b) ATL for 10 users attending the copy phrase task in RSVP Keyboard experiment.}}
        \label{fig:bciTyping}
\end{figure}
}

\newcommand{\speedAccQuery}{
 \begin{figure*}[t!]
       \centering
        \subfloat[]{\label{fig:stopping}\includegraphics[width=0.45\textwidth, trim=0in 0in 0in 0.2in, clip =False]{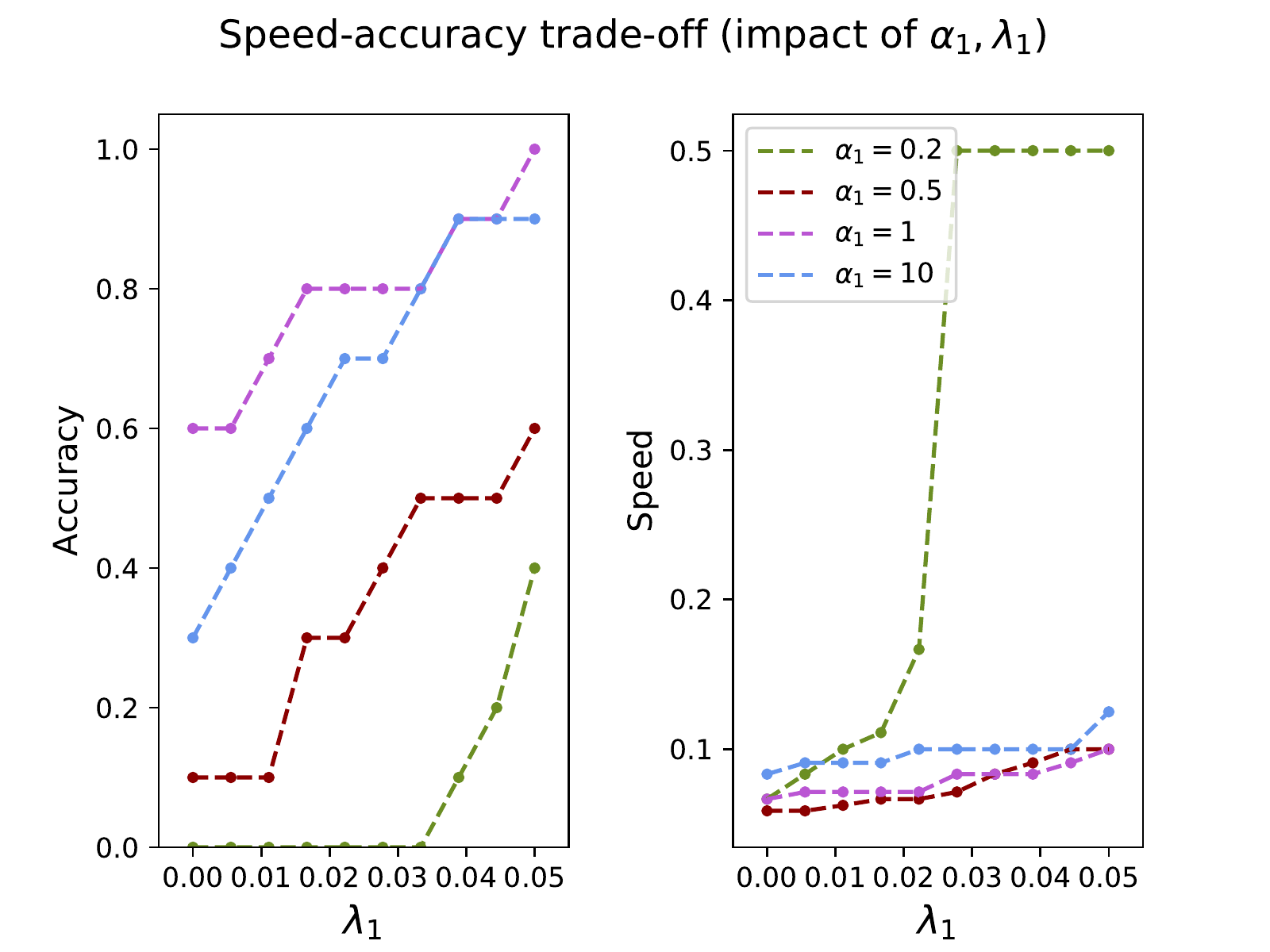}}
        \subfloat[]{\label{fig:querying}\includegraphics[width=0.45\textwidth, trim=0in 0in 0in 0.2in, clip =False]{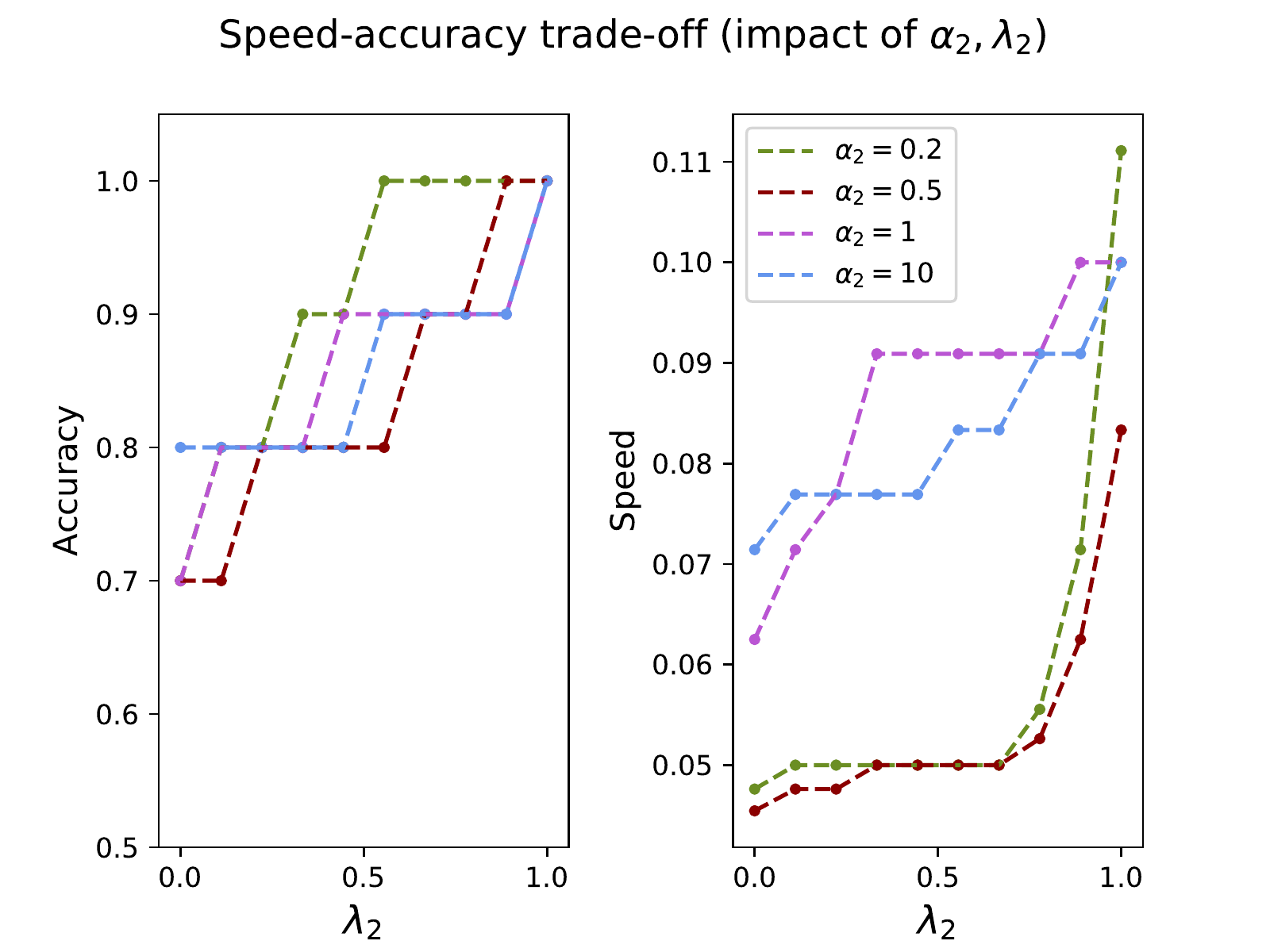}}
        \caption{{Impact of $\alpha$ and $\lambda$ on the speed and accuracy of inference in the proposed RBI framework. (a) Given $\alpha_2 , \lambda_2=1$, speed-accuracy trade off is computed as a function of $\alpha_1$ and $\lambda_1$. (b) Using $\alpha_1=\infty$ and $\lambda_1=0$, the accuracy and speed (inverse of the number of sequences) were calculated as a function $\alpha_2$ and $\lambda_2$.}}
        \label{fig:speedAccQuery}
    \end{figure*}
}

\newcommand{\decsionBoundary}{
\begin{figure}[b!]
\centering
\includegraphics[scale=0.85]{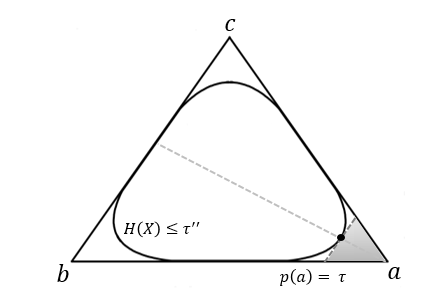} 
\caption{Feasible sets in inference problem for two decision boundaries: maximum of the posterior and entropy.}
\label{fig:decsionBoundary}
\end{figure}
}

\newcommand{\diffRenyiEntropy}{
 \begin{figure*}[t!]
      \centering
      \subfloat[]{\includegraphics[width=0.35\textwidth]{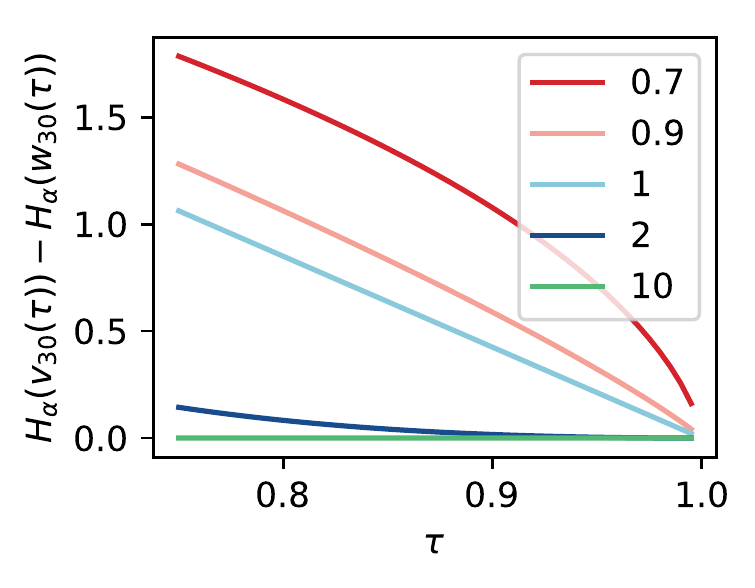}}
       \subfloat[]{\includegraphics[width=0.35\textwidth]{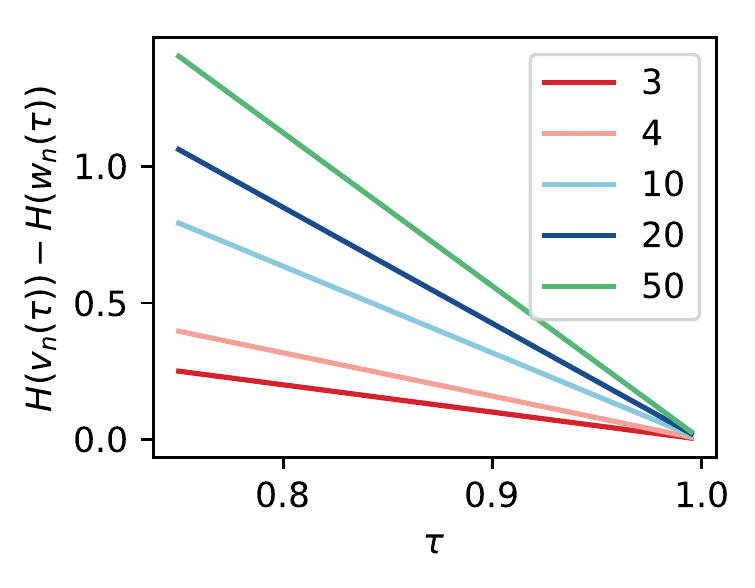}}
        \caption{Difference between entropy values of the probabilities $v_n,w_n$ for a given maximum value $\tau$. (a) Presents the entropy differences with differing $\alpha$; (b) presents the entropy differences with differing $n$.}
\label{fig:diffRenyiEntropy}       
\end{figure*}
}

\newcommand{\diffRenyiAlpha}{
\begin{figure*}[t!]
      \centering
      \subfloat[]{\includegraphics[width=0.35\textwidth]{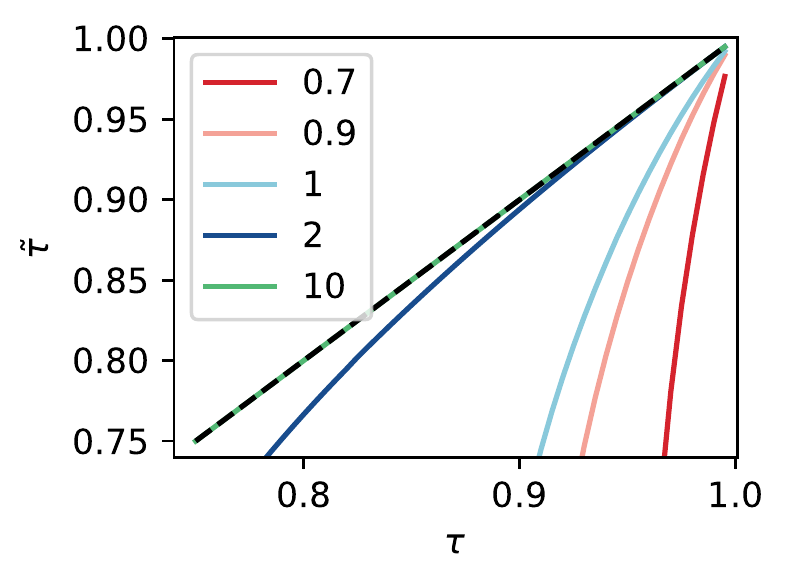}}
        \subfloat[]{\includegraphics[width=0.35\textwidth]{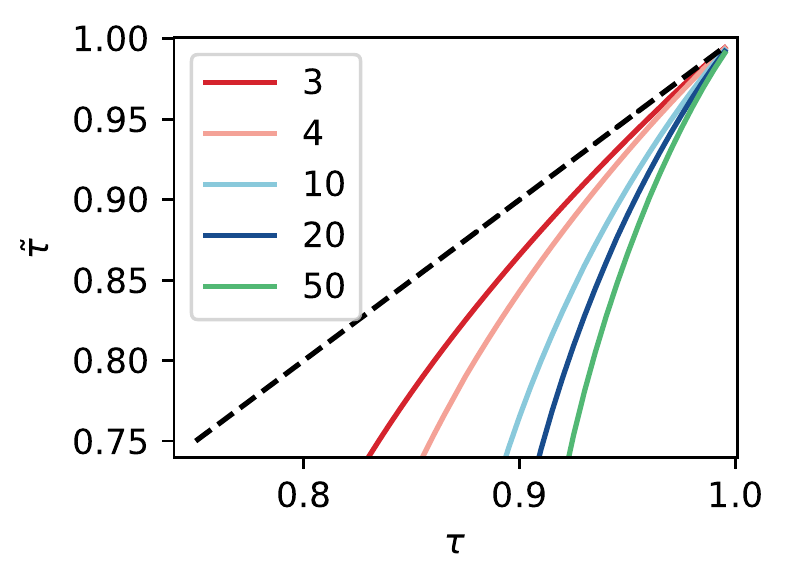}}
        \caption{(a) Presents the corresponding $(\tau,\tilde{\tau})$ tuples where $n=30$ for different $\alpha$ values; (b) presents the corresponding $(\tau,\tilde{\tau})$ tuples where $\alpha=1$ for different alpha $n$ values.}
\label{fig:diffRenyiAlpha}
\end{figure*}
}

\newcommand{\simplexSimulations}{
\begin{figure*}[t!]
      \centering
      \subfloat[]{\label{fig:simplexquery}\includegraphics[width=0.6\textwidth]{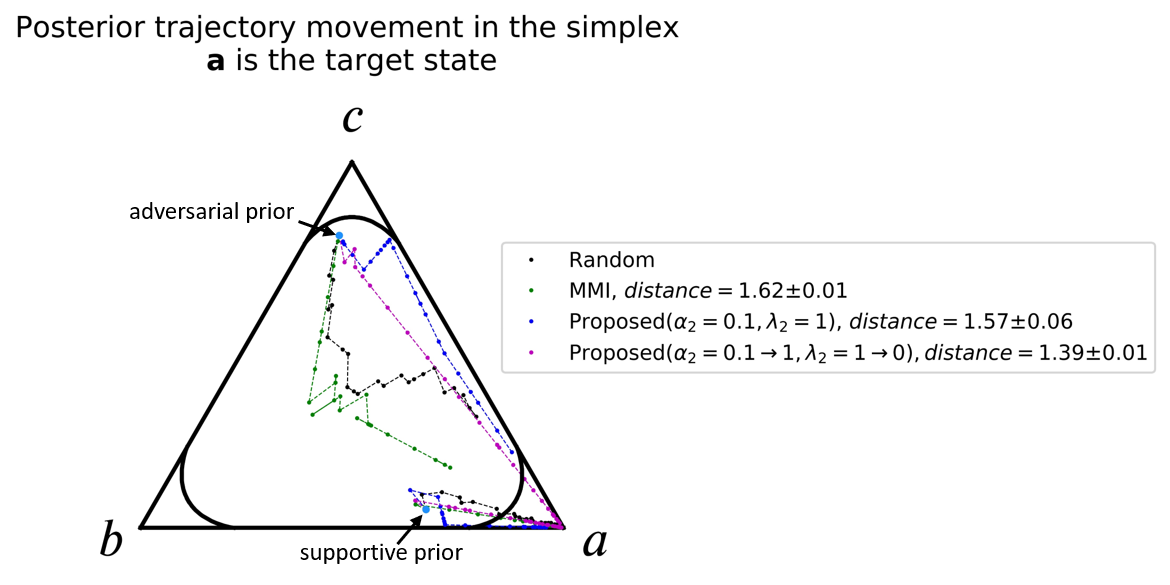}}
        \subfloat[]{\label{fig:SimplexSpeed}\includegraphics[width=0.35\textwidth]{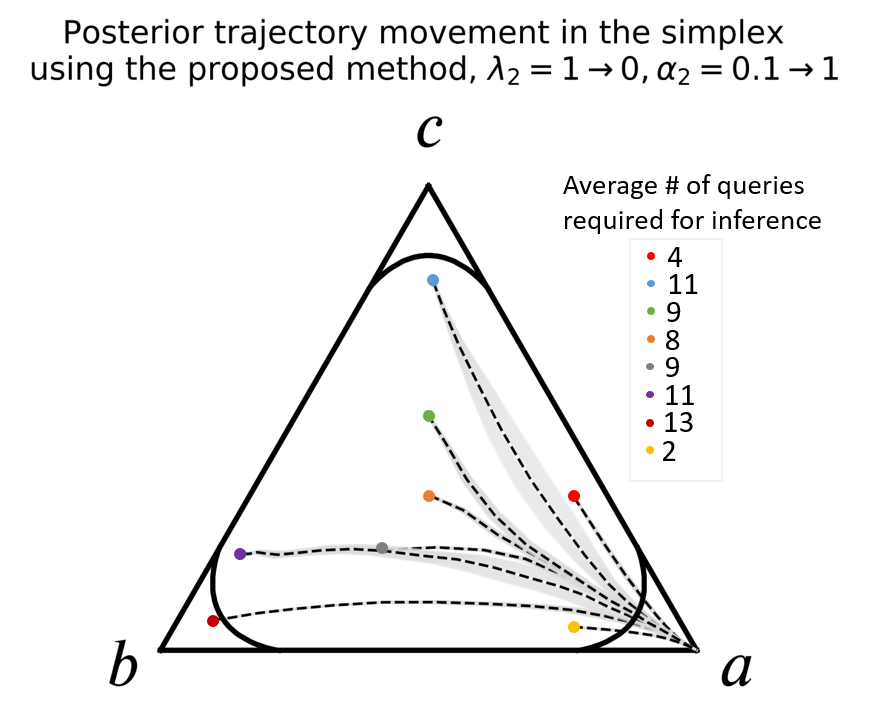}}
        \caption{ Geometric representation of the RBI problem in probability simplex for three elements. (a) Recursive posterior transition in the simplex using Random, MMI, and the proposed query methods. (b) The effect of the prior on the posterior movement trajectories in the simplex. Each blue dot represents a prior. The reported average trajectory and its variability computed for 500 Monte-Carlo simulations. For all simulations, two 1D Gaussian distributions, $\mathcal{N}(0, 1)$, $\mathcal{N}(3, 1.5)$ are used to model target and non-target distributions.}
        \label{fig:simplexSimulations}
\end{figure*}
}
\newcommand{\resultsTable}{
\begin{table*}[h]
\caption{Performance of the proposed framework using different setting of parameters in the restaurant recommender system, for two types of diners for 80 restaurants' reviews. ACC presents the accuracy of the estimation and speed is inverse number of sequences. MMI is equivalent to the proposed method using $(\alpha_1:\infty, \alpha_2:1, \lambda_1:0, \lambda_2:0)$, Proposed-c$\alpha$ is equivalent to the proposed method using $(\alpha_{1,2}:1, \lambda_1:0 	\rightarrow 0.01, \lambda_2:1 \rightarrow 0)$ with constant $\alpha$ values, and Proposed-v$\alpha$ is equivalent to the proposed method using $(\alpha_{1,2}: 0.1 	\rightarrow 1, \lambda_1: 0 \rightarrow 0.01, \lambda_2: 1 \rightarrow 0)$, where $\lambda_{1,2}$, in which both $\alpha$ and $\lambda$ are variable and being changed over sequences.}
\label{table:resultsTable}
\centering
{\begin{tabular*}{38pc}{@{\extracolsep{\fill}}ccccccccc@{}}
   \toprule
    &\multicolumn{2}{c}{Random}
    &
    \multicolumn{2}{c}{MMI}
    &
    \multicolumn{2}{c}{Proposed-c$\alpha$}
    &
    \multicolumn{2}{c}{Proposed-v$\alpha$}
    \\\cmidrule(r){2-3}\cmidrule(l){4-5}\cmidrule(r){6-7} \cmidrule(r){8-9}
    \small{Case} & \small{ACC} & \small{Speed} & \small{ACC} & \small{Speed} & \small{ACC} & \small{Speed} & \small{ACC} & \small{Speed}\\
    \hline
    Conventional Diner & 85.29 & 0.0500 & 92.03 & 0.1400 & 92.10 & 0.1400 & 92.51 & 0.1600 \\ [.02in]
    Adversarial Diner  & 57.88 & 0.0104 & 74.57 & 0.0172 & 75.72 & 0.0185 & \textbf{80.73} & \textbf{0.0189} \\ [.02in]
    \bottomrule
\end{tabular*}}{}
\end{table*}
}
\def\infinity{\rotatebox{90}{8}}

\begin{document}

\title{Active recursive Bayesian inference using Rényi information measures}
\author{*Yeganeh M. Marghi{$^1$}, *Aziz Ko\c{c}anao\u{g}ullar\i{$^1$}, Murat Ak\c{c}akaya{$^2$}, Deniz Erdo\u{g}mu\c{s}{$^1$} \newline
  %\IEEEauthorblockA{\IEEEauthorrefmark{2} Northeastern University}\\
  %\IEEEauthorblockA{\IEEEauthorrefmark{3} University of Pittsburgh}
  {{$^1$}Northeastern University, {$^2$}University of Pittsburgh}
\thanks{* Joint first authors.}
\thanks{This paper is under consideration at Pattern
Recognition  Letters.}
%\thanks{This work is supported by NSF (IIS-1149570, CNS-1544895, IIS-1715858, IIS-1717654, IIS-1844885, IIS-1915083, IIS-1854827), DHHS (90RE5017-02-01), and NIH (R01DC009834).}% <-this % stops a space
%\thanks{Yeganeh M Marghi, Aziz Ko\c{c}anao\u{g}ullar\i, and Deniz Erdo\u{g}mu\c{s} are with the Cognitive Systems laboratory (CSL), Electrical and Computer Engineering Department, Northeastern University, Boston, MA, USA.}
%\thanks{Murat Ak\c{c}akaya is with the Department of Electrical and Computer Engineering, University of Pittsburgh, PA 15261, USA.}
}
\maketitle
\begin{abstract}
Recursive Bayesian inference (RBI) provides optimal Bayesian latent variable estimates in real-time settings with streaming noisy observations. Active RBI attempts to effectively select queries that lead to more informative observations to rapidly reduce uncertainty until a confident decision is made. However, typically the optimality objectives of inference and query mechanisms are not jointly selected. Furthermore, conventional active querying methods stagger due to misleading prior information. Motivated by information theoretic approaches, we propose an active RBI framework with unified inference and query selection steps through Rényi entropy and $\alpha$-divergence. We also propose a new objective based on Rényi entropy and its changes called Momentum that encourages exploration for misleading prior cases. The proposed active RBI framework is applied to the trajectory of the posterior changes in the probability simplex that provides a coordinated active querying and decision making with specified confidence. Under certain assumptions, we analytically demonstrate that the proposed approach outperforms conventional methods such as mutual information by allowing the selections of unlikely events. We present empirical and experimental performance evaluations on two applications: restaurant recommendation and brain-computer interface (BCI) typing systems.
\end{abstract}
\section{Introduction}
\label{sec:Introduction}
% Method
Recursive Bayesian inference (RBI) is a general probabilistic framework to estimate the unknown probability distribution of latent states through a recursive querying process over time. Due to advantages of Bayesian approaches, for state estimation, it is also preferred to fuse the obtained observation with a prior information in a Bayesian manner. Accordingly, the state estimation and the estimation confidence depend on the posterior distribution recursively calculated over the state. Since, the observations are noisy, to achieve a high confidence and to decrease ambiguity in the state estimation, the system probes the environment through multiple recursions of queries~\cite{bergman1999recursive, sarkka2013bayesian}. Throughout this manuscript, we denote each recursion and corresponding confidence update as a \textit{sequence}.

The RBI framework consists of two main objectives: evidence collection (potentially through active querying through a sequence of queries) and inference. The goal of the inference is to estimate latent state that is the most likely to be the intended state. The inference objective is defined based on a Bayes decision policy. Typically the \textit{maximum a posteriori} (MAP) estimation is used to minimize zero-one risk. Decision is constrained with a pre-defined confidence level over the posterior distribution  to prevent premature decisions. To reach the confidence level, the system attempts to make inference after observing the evidence based on the queries. However, the evidence collection  process is costly, especially when the evidence is not collected through carefully selected queries. Therefore, query optimization are required to make the recursive inference more practical and efficient. 

From a state-estimation perspective, query optimization in the recursive state estimation is often performed by greedy selection, which can be split into three subgroups of 1) expected posterior maximization~\cite{wilson2012bayesian}, in which the Bayesian model has been used for the querying process; 2) Fisher information-based approaches~\cite{hoi2006batch,chepuri2015sparsity,sourati2017probabilistic}, and 3) information theory-based approaches such as entropy minimization or maximum mutual information (MMI)~\cite{golovin2010near,higger2017recursive,jedynak2012twenty,MoghadamJSTS,mainsahBCI2018}. It is shown that all these approaches for optimum sequence design through query selection lead to the selection of \textit{N-best} queries with respect to the current belief~\cite{koccanaogullari2018analysis,tsiligkaridis2014collaborative}. As an example of posterior maximization, Wilson et al. in~\cite{wilson2012bayesian}, proposed a Bayesian query optimization to elicit the target with as few queries as possible using a sampling with a Monte Carlo estimate dependent variational distance measure. Using Fisher information, Sourati et al. ~\cite{sourati2017probabilistic} introduced a Fisher information-based ratio for an active sample selection for training a classifier with tractable computation. The proposed ratio shifts the classifier parameters towards the most informative point in the parameter space. Since these parameters were essentially obtained through a posterior maximization, this approach can also be adopted to maximize the information of the argument in the posterior. Hierarchical sampling proposed by Dasgupta and Hsu~\cite{dasgupta2008hierarchical} is another active learning framework proposed for query selection, which forms a query subset partitioned with different objectives.

All these approaches for optimum sequence design through query selection would lead to the selection of \textit{N-best} queries decided based on the probability distribution defined over the latent states, under two assumptions: (i) unimodality of observation distribution and (ii) finite subset of queries~\cite{koccanaogullari2018analysis, tsiligkaridis2014collaborative}. \textit{N-best} selection queries based only on the latest belief over the states may not always provide the best performance, especially at the beginning of the RBI process, as they always exploit the current knowledge and lack further exploration beyond current belief.  

As an example, consider a recommender system. In such a system, the current belief is always influenced by the history of the users' profiles. Accordingly, by employing the \textit{N-best} selection, the system recommends items similar to the ones selected by peers, or reminiscent of the ones the user has previously showed interest in. However, these systems are not as effective, when the user is looking for a new item, e.g. in the case of an adventurous diner. In this case, the prior information behaves in an adversarial manner, causing a longer estimation procedure, or may lead to an unsatisfactory estimation due to the limited number of observations. Therefore, instead of exploiting the current belief, such systems can benefit from exploration of evidence beyond the current belief.  
%It would be beneficial to investigate a trade-off between exploration and exploitation in RBI problem.
%\newline

In addition to the sequence design based on query selection, in an RBI framework, inference is usually constrained with a posterior threshold. Such a threshold increases the number of query sequences to obtain enough evidence for confident inference. Therefore, in RBI there is a trade-off between speed and accuracy. Here, we propose a unified active RBI framework to balance the speed and accuracy trade-off and enhance RBI performance. Our framework is particularly effective in adversarial scenarios where a balance between exploitation and exploration needs to be achieved.The proposed framework utilizes the Bayesian approach for updating the probability distribution defined over the states using posterior trajectory to enhance speed accuracy trade-off in state estimation. By defining query selection, inference objectives and confidence constraints based on Rényi entropy, our mathematical formulation unifies the query selection and inference steps of RBI. Such a formulation enables us to analytically demonstrate that the proposed policy for query selection that finds a compromise between exploitation and exploration performs at least as good as the methods that make \textit{N-best} selections relying on the exploitation of the current belief. Moreover this formulation enables us to formulate the subset query selection optimization problem as a tractable problem~\cite{baram2004online}. Finally, we provide analytical  results to show that the proposed unified approach has theoretical guarantees to outperform the methods based only on mutual information or entropy.

%\textcolor{blue}{, in which the problem can be solved in polynomial time (in the size of the input)}
%The essence of the Bayesian approach in this framework is to provide a mathematical formulation describing how the probability value of a state should change its probability at a given recursion (current belief) in presence of new evidence. We formulate the query selection and decision making criterion objectives based on a general formulation of information measures, Rényi measures.
%
In summary, our novel contributions are (i) unifying the inference and query selection steps of RBI through Rényi entropy; (ii) proposing a new objective based on Rényi entropy and changes in the Rényi entropy, i.e. \textit{Momentum} that encourages exploration for adversarial cases,  (iii) proposing a new stopping criterion for decision making, (iv) providing a short-term policy evaluation to demonstrate the proposed method has theoretical performance guarantees under reasonable assumptions, and (v) presenting a novel geometrical representation for RBI problem that allows us to investigate how the query optimization influences the estimation process and how the posterior belief changes over sequences in the probability \textit{simplex}. To illustrate the performance of the proposed method, we consider two testbeds, (a) restaurant recommendation system, and (b) a brain-computer interface (BCI)-based typing system.  

\section{Preliminaries}
\label{sec:Preliminaries}
% Preliminary
In the recursive state estimation, the learner (system) is tasked to update its belief of the distribution over state candidates by collecting observations as a result of querying the environment. The state is denoted by $\sigma$. We assume that the state is an element of a finite set denoted by $\mathcal{A}$. The learner proceeds with the evidence collection for state estimation  %with the estimation through a sequential decision making process containing 
through \textit{sequences of queries} indexed by $s$ %including 
which is divided into $N$ \text{trials} indexed by $i$. The system decides on a subset of queries $\Phi_s \triangleq \{\phi^{1}_{s}, \dots, \phi^{N}_{s}\}$ at the beginning of each sequence, where $\phi^i_s \in \mathcal{Q}$ and $\mathcal{Q}$ is the set of queries. After system queries the environment, corresponding evidence $\beps_s \triangleq \{\eps^{1}_{s}, \dots, \eps^{N}_{s}\}$ is observed . Due to the noise in an observation, estimation of $\sigma$ requires recursion of multiple sequences. This process can be well formulated as a RBI problem. By formulating the problems in a Bayesian estimation framework, we decompose the estimation into inference $(I)$ and querying $(Q)$ objectives, which is expressed as follows.
\begin{equation}
    \label{eq:originProb}
    \begin{split}
    (I): \ &\hat{\sigma} = \displaystyle{\arg\max_{\sigma}} f_I(\sigma,\mathcal{H}_s) \\ 
    &\mbox{ s.t.} \ g_I(\sigma,\mathcal{H}_s)>0 \\[.15in]
    (Q): \  &{\Phi}_{s+1} = \displaystyle{\arg\max_{\Phi}} f_Q(\sigma,\Phi,\mathcal{H}_{s}) 
    \end{split}
\end{equation}
Here, $\mathcal{H}_{s} \triangleq \{\beps_{1:s}, \Phi_{1:s}, \mathcal{H}_0\}$ represents the task history and $\mathcal{H}_0$ represents the prior information. Where $f$ and $g$ are denoting arbitrary functions to distinctly represent objectives and the constraint respectively. The system alternates between $(I)$ and $(Q)$ to collect evidence until the optimality objective measure, i.e. an equality constraint $g_I$ is satisfied. Accordingly, the choice of constraint and query optimization objective is critical to reduce the cost e.g., number of queries.
% Add figure - graphical model
\graphicalModel

To represent the probabilistic relationship among variables, the graphical model of the RBI problem is illustrated in Figure~\ref{fig:graphicalModel}. In this figure, $\bell_s$ is the label for each observation at sequence $s$. In this paper, we assume observations are independent of each other given the label sequence, therefore the posterior probability is defined as the following;
\begin{equation*}
\label{eq:probModel_PGM}
    \begin{array}{rl}
        p(\sigma | \beps_{s},\Phi_{s}, \mathcal{H}_{s-1})=p(\sigma|\mathcal{H}_0)\displaystyle\prod_{j=1}^s\frac{p(\beps_j | \sigma, \Phi_j)}{p(\beps_j | \Phi_j, \mathcal{H}_{j-1})}
    \end{array}
\end{equation*}
where $p(\beps_j | \sigma, \Phi_j)$ and $p(\beps_j | \Phi_j, \mathcal{H}_{j-1})$ can be obtained trough marginalization over $\bell$ as,
\begin{equation*}
\label{eq:probModel_PGM}
    \begin{array}{rl}
        p(\beps_j | \sigma, \Phi_j)
        &=\displaystyle{\sum_{\bell_j}} \ p(\beps_j | \bell_j)p(\bell_j | \sigma, \Phi_j) \\
        p(\beps_j | \Phi_j, \mathcal{H}_{j-1})&=\displaystyle{\sum_{\bell_j}} \ p(\beps_j | \bell_j)p(\bell_j | \Phi_j, \mathcal{H}_{j-1})
    \end{array}
\end{equation*}
%
%\begin{equation*}
%\label{eq:probModel_PGM}
%    \begin{array}{rl}
%        p(\sigma | \beps_{s},\Phi_{s}, \mathcal{H}_{s-1})
%        =p(\sigma|\mathcal{H}_0)\displaystyle\prod_{j=1}^%s\frac{\displaystyle{\sum_{\bell_j}}p(\beps_j | %\bell_j)p(\bell_j | \sigma, %\Phi_j)}{\displaystyle{\sum_{\bell_j}}p(\beps_j | %\bell_j)p(\bell_j | \Phi_j, \mathcal{H}_{j-1})}
%    \end{array}
%\end{equation*}
Accordingly, we can have:
\begin{equation*}
\label{eq:label_probmodel}
    \begin{array}{rl}
       p(\sigma | \bell_j, \Phi_j, \mathcal{H}_{j-1}) =\displaystyle{\frac{p(\bell_j | \sigma, \Phi_j)p(\sigma |\mathcal{H}_{j-1})}{p(\bell_j | \Phi_j, \mathcal{H}_{j-1})}} \\[.2in] =\displaystyle{\frac{p(\bell_j | \sigma, \Phi_j)p(\sigma |\mathcal{H}_{j-1})}{\displaystyle{\sum_{v \in \mathcal{A}}}p(\bell_j | v, \Phi_j)p(v |\mathcal{H}_{j-1})}}
    \end{array}
\end{equation*}
where $p(\bell_j | \sigma, \Phi_j)$ expresses the probabilistic relationship between each subset of queries and a state, which is typically learned by \textit{query understanding}
techniques~\cite{liu2013query}. Since Query understanding is out of the scope of this study, it is assumed that $p(\bell_j | \sigma, \Phi_j)$ is given as an external input.
%By query understanding, we refer to the process of analysing the information obtained from a query and transferring that from the query domain $\mathcal{Q}$ to the state space $\mathcal{A}$. Depending on the application, there exits several techniques to analyze the query content~\cite{liu2013query}. Query understanding is out of the scope of this study and it is assumed that $p(\bell_j | \sigma, \Phi_j)$ is given as an external input (\textcolor{red}{Give an example here to be more specific. This is currently very vague}).
%

Conventionally, in the active RBI framework, $f_I$ is a function of posterior probability and changes in probability domain and state constraint is a confidence level threshold $\tau \in \mathbb{R}$ on the state posterior probability. Moreover, query selection is typically achieved through information theory-based techniques such as MMI~\cite{jedynak2012twenty}, which is equivalent to conditional entropy minimization when $\beps_s$ corresponding to the upcoming sequence is not observed. Accordingly, $f_I$ and $f_Q$ objectives and $g_I$ are defined as the following;
\begin{equation}
    \label{eq:convMIform}
    \begin{array}{ll}
    (I): &\hat{\sigma} =  \displaystyle{\arg\max_{\sigma\in\mathcal{A}}} \  p(\sigma| \mathcal{H}_s) \\[.1in]
    &\text{s.t.} \ p({\sigma}| \mathcal{H}_s) > \tau \\[.15in]
    (Q): \  &{\Phi}_{s+1} = \displaystyle{\arg\max_{\Phi\in\mathcal{Q}^N}} \ -H(\sigma | \beps_{s}, \Phi, \mathcal{H}_s)
    \end{array}
\end{equation}
Since $-H(\sigma | \beps_{s}, \Phi, \mathcal{H}_s)$ is a submodular and non-increasing set function, the sequential subset query selection can be achieved via a greedy approach by optimizing $f_Q$ for each query with theoretical guarantees~\cite{barron1986entropy,johnson2004fisher,farias2011irrevocable}. Therefore, query selection objective can be reformulated for each query as the following;
\begin{equation}    
    \label{eq:action_value_mmi}
   \begin{array}{ll}
    (Q): \  &\phi^i_{s+1} = \arg\max\limits_{\phi} \ -H(\sigma| \eps^i_{s+1} , \phi, \mathcal{H}_{s}).
   \end{array}
\end{equation}
%
%
% Method
\section{Integrating Inference and Querying Objectives}
\label{sec:Renyi}
A known challenge in optimizing multi-objective functions such as~\eqref{eq:originProb}, is that each individual objective demands its own evaluation criteria and constraints. Therefore, the optimal solution for one objective may not be optimum for the other(s). For instance, the optimal query obtained by maximizing $f_Q$ in~\eqref{eq:originProb} may not be the optimal query to be used in $f_I$ when estimating the target state $\sigma$. Accordingly, it is ideal to use a unified objective function, which conjoins $(Q)$ and $(I)$ by unifying $f_Q$ and $g_I$, such that optimizing $f_Q$ results in a query satisfying $g_I$.
In conventional representation of the objectives as defined in \eqref{eq:convMIform}, $(Q)$ is designed to select subset of queries to reduce the ambiguity over the state estimates whereas the $(I)$, which is constrained with a posterior threshold. We show the unity between these two identities using Rényi entropy~\cite{renyi1961measures,song2001renyi,li2016renyi}. Accordingly, the active RBI framework can be represented as,
\begin{equation}
    \label{eq:Renyi_RBI}
    \begin{array}{ll}
    (I): &{\sigma} =  \displaystyle{\arg\max_{\sigma \in \mathcal{A}}} \  p(\sigma| \mathcal{H}_s) \\[.1in]
    &\text{s.t.} \ H_{\alpha_1}({\sigma}| \mathcal{H}_s) < \tau' \\[.15in]
    (Q): \  &{\phi}^{i}_{s+1} = \displaystyle{\arg\max_{\phi \in \mathcal{Q}}} \ -H_{\alpha_2}({\sigma}| \eps^i_{s+1}, \phi, \mathcal{H}_s) \\[.15in]
    \end{array}
\end{equation}
where, $\tau'$ is the upper bound of the acceptable entropy, $\alpha_1 , \alpha_2 \in [0, \infinity)$ are the orders of entropy measures for $(I)$ and $(Q)$ respectively. $\alpha$-entropy and conditional entropy are defined as the following;
\begin{equation}
    \label{eq:Renyi_entropy}
    \begin{array}{rl}
         H_{\alpha_{1}}(\sigma|\mathcal{H}_s) = \displaystyle{\frac{1}{1-\alpha_1}}\log{\left(\displaystyle{\sum_{\sigma \in \mathcal{A}} \ p^{\alpha_1}(\sigma|\mathcal{H}_s)} \right)}
        \end{array}
\end{equation}
\begin{equation}
    \label{eq:Renyi_conentropy}
    \begin{split}
         H_{\alpha_{2}}(\sigma|\beps, \phi, \mathcal{H}_s) = \displaystyle{\frac{1}{1-\alpha_2}}\mathbf{E}_{p(\eps|\phi,\mathcal{H}_s)} \left(\vphantom{\sum_{\sigma \in \mathcal{A}} \ p^{\alpha}(\sigma|\beps, \phi, \mathcal{H}_s)}\right. \hspace{1.3cm}\\
         \left.\log{\left(\displaystyle{\sum_{\sigma \in \mathcal{A}} \ p^{\alpha_2}(\sigma|\beps, \phi, \mathcal{H}_s)} \right)}\right)
    \end{split}
\end{equation}
Conclusively, for $\alpha_1=\infinity$ and $\alpha_2=1$ the objectives in \eqref{eq:convMIform} and \eqref{eq:Renyi_RBI} are identical. Therefore, $\alpha_1$ and $\alpha_2$ are tuning hyperparameters that adjust a balance between the maximum posterior condition ($\alpha_1=\infinity$) and Shannon entropy ($\alpha_2=1$). Moreover, since $H_{\alpha_2}$ is also a submodular and non-increasing set function as shown in \cite{madiman2008entropy,van2010renyi}, the sequential subset query selection in \eqref{eq:Renyi_RBI} is performed via a greedy algorithm. 
\section{A Posterior Changes-based Objective Function}
\label{sec:An Objective function of Posterior Changes Trajectory}
The proposed framework in (\ref{eq:Renyi_RBI}) is using a measure of expected information gain with the current belief (posterior) of an estimate and results in the MMI selection.
However, greedy selections based on posterior might initially suffer if the prior information is misleading. In contrast to prior belief, if the system state is one of the least probable scenarios (according to prior), i.e. the adversarial case, the system needs to overcome the misleading prior via exploration. 

To encourage exploration in the query selection $(Q)$, previously we have proposed a new objective called \textit{Momentum}, which is a function of posterior changes over sequences~\cite{koccanaougullari2018optimal}. We have shown that including Momentum in $f_Q$ provides a trade-off between exploration and exploitation and compared to the \textit{N-best} method, it increases the chance of the unlikely queries to be selected. In~\cite{marghi2019history}, we have also demonstrated that the accuracy/speed ratio can be improved by adding the Momentum term in the stopping criterion, in a RBI problem. While the previously introduced Momentum function has shown promising results, it cannot be utilized in the integrated RBI framework using a general class of information measures in~\eqref{eq:Renyi_RBI}. 
To extend the existing methods, in this study, we redefine the Momentum objective by proposing a general family for Momentum that can be used for both query selection and stopping criterion in the unified RBI framework.
Accordingly, the alternative $f_{Q}$ for \eqref{eq:Renyi_RBI} is expressed as the following;
\begin{equation}
    \label{eq:Renyi_monetum_query}
    \begin{array}{ll}
    {\phi}^{i}_{s+1} = \displaystyle{\arg\max_{\phi \in \mathcal{Q}}} \ -H_{\alpha_2}({\sigma}| \eps^i_{s+1}, \phi, \mathcal{H}_s) + \lambda_1 M_{\alpha_2}(\phi,\mathcal{H}_s) \\[.15in]
    \end{array}
\end{equation}
\begin{equation}    
        \label{eq:ave_momentum_query}
        \begin{array}{ll}
        M_{\alpha}(\phi , \mathcal{H}_{s}) = \displaystyle\frac{1}{s}\displaystyle\sum_{n=0}^{s-1}  m_{\alpha}(\phi , \mathcal{H}_{n}) \\
        = \displaystyle\frac{1}{s}\displaystyle\sum_{n=0}^{s-1} \displaystyle{\frac{1}{(\alpha - 1)}}\log{\displaystyle{\sum_{\sigma \in \mathcal{A}}}\displaystyle{\frac{p^{\alpha}(\sigma|\eps, \phi, \mathcal{H}_{n})}{p^{{\alpha}-1}(\sigma|\mathcal{H}_{n})}}p(l=1|\sigma,\phi)}
        \end{array}
    \end{equation}
where $m_{\alpha}(\phi, \mathcal{H}_{n})$ is called \textit{Q-Momentum}, $M_{\alpha_2}(\phi,\mathcal{H}_s)$ is the average Q-Momentum and $\lambda_1$ is the tuning parameter between objectives. Momentum term by definition is a state-directed average of $\log$-posterior changes over all previous sequences which allows for a one-step advanced approximation of overall conditional entropy. 

Likewise, inclusion of Momentum in $(I)$ can sped up the estimation process through constraint relaxation in~\eqref{eq:Renyi_RBI} as follows.
\begin{equation}    
        \label{eq:ave_momentum_stop}
        \begin{array}{ll}
        \mathcal{D}_{\alpha}(\mathcal{H}_{s}) = \displaystyle\frac{1}{s}\displaystyle\sum_{n=0}^{s-1}  d_{\alpha}(\mathcal{H}_{n}) \\
        = \displaystyle\frac{1}{s}\displaystyle\sum_{n=0}^{s-1} \displaystyle{\frac{1}{(\alpha - 1)}}\log{\displaystyle{\sum_{\sigma \in \mathcal{A}}}\displaystyle{\frac{p^{\alpha}(\sigma|\mathcal{H}_{n+1})}{p^{{\alpha}-1}(\sigma|\mathcal{H}_{n})}}}
        \end{array}
\end{equation}
where $d_{\alpha}(\mathcal{H}_{n})$ is called \textit{I-Momentum} and $\mathcal{D}_{\alpha}(\mathcal{H}_s)$ is the average I-Momentum.
Accordingly, the unified framework for RBI problem can be expressed as;
\begin{equation}
    \label{eq:Renyi_monetum_RBI}
    \begin{array}{ll}
    (I): &{\sigma} =  \displaystyle{\arg\max_{\sigma \in \mathcal{A}}} \  p(\sigma| \mathcal{H}_s) \\[.1in]
    &\text{s.t.} \ H_{\alpha_1}({\sigma}| \mathcal{H}_s) - \lambda_1 \mathcal{D}_{\alpha_1}(\mathcal{H}_s) < \tau' \\[.15in]
    (Q): &{\phi}^{i}_{s+1} = \displaystyle{\arg\max_{\phi \in \mathcal{Q}}}  -H_{\alpha_2}({\sigma}| \eps^i_{s+1}, \phi, \mathcal{H}_s) + \\
    & \hspace{1.1in} \lambda_2 M_{\alpha_2}(\phi,\mathcal{H}_s) \\[.15in]
    \end{array}
\end{equation}
For a single query $\phi$ asked in the RBI framework, the posterior probability is updated according to \textit{Bayes rule}, which implies that at each sequence, the posterior change is equivalent to $p(\sigma | \mathcal{H}_{s-1})\left[\displaystyle{\frac{p(\eps|\sigma, \phi)}{p(\beps|\phi)}}- 1\right]$. Accordingly, the posterior change is a function of $\displaystyle{\frac{p(\eps|\sigma, \phi)}{p(\eps|\phi)}}$, called \textit{likelihood evidence ratio} and based on the Bayes rule, it is equivalent to $\displaystyle{\frac{p(\sigma|\mathcal{H}_s)}{p(\sigma|\mathcal{H}_{s-1})}}$. In fact, both introduced Momentum terms in $(I)$ and $(Q)$ are function of posterior changes. In the following sections, we will show how Momentum enhances the speed and accuracy of the active RBI process.

Pseudocode of the proposed unified active RBI can be found in Algorithm~\ref{alg:Active Inference for Decision Making}.
System loops querying until the predefined stopping criterion is satisfied. At sequence $s+1$, $N$ number of unique queries are selected from the query set $\mathcal{Q}$. After the query subset is selected, the corresponding evidence $\beps_{s+1}$ is observed as a response to the queries. Based on the observation, system updates the belief over the states and current stopping condition. Once a feasible solution arises, system returns the candidate with maximum of the current posterior. The initial \textit{stop} criterion is adjusted such that the system is required to query at least once to avoid instant termination.
%
% algorithm
\begin{algorithm}[!t]
\label{alg:proposed_greedy}
\caption{Greedy approach for the active recursive Bayesian inference framework}
\label{alg:Active Inference for Decision Making}
\begin{algorithmic}[1]
\State \textbf{initialize} $\mathcal{A},\mathcal{Q},\alpha_1\in[0,\infty) ,\alpha_2\in[0,\infty),\tau',\lambda_1,\lambda_2$
\State $s\gets 0,\mathcal{H_s}\gets \lbrace \mathcal{H}_0 \rbrace$
\State $\text{stop} \gets \tau' + 1 $
\While{ $\text{stop} > \tau' $}
\State $\mathcal{Q}'\gets \mathcal{Q}$
\For{ $i \in \lbrace 0,1,\cdots,N \rbrace$} %\Comment{\small{Batch query selection}}
    \State ${\phi}^{i}_{s+1} \gets \displaystyle{\arg\max_{\phi \in \mathcal{Q}'}} \small{-H_{\alpha_2}({\sigma}| \eps^i_{s+1}, \phi, \mathcal{H}_s) +
    \lambda_2 M_{\alpha_2}(\phi,\mathcal{H}_s)}$
    \State $\mathcal{Q}' \gets \mathcal{Q}' \setminus \lbrace {\phi}^{i}_{s+1}\rbrace$
\EndFor
\State $\varepsilon_{s+1} $ observe for $\Phi_{s+1}$ \Comment{\small{Evidence Collection}}
\State $\mathcal{H}_{s+1} \gets \mathcal{H}_{s} \cup \lbrace \varepsilon_{s+1}, \phi_{s+1} \rbrace$ \Comment{\small{Update History}}
\State $s\gets s+1$
\State $\text{stop}\gets H_{\alpha_1}({\sigma}| \mathcal{H}_s) - \lambda_1 \mathcal{D}_{\alpha_1}(\mathcal{H}_s)$ \Comment{\small{\eqref{eq:Renyi_monetum_RBI}}}
\EndWhile
\State $\hat{{\sigma}} =  \displaystyle{\arg\max_{\sigma \in \mathcal{A}}} \ p(\sigma| \mathcal{H}_s)$
\State \textbf{return} $\hat{\sigma}$
\end{algorithmic}
\end{algorithm}
\subsection*{Analytical Evaluation} 
\label{subsec:Analytical Evaluation}
The main purpose of Q-Momentum objective in the proposed active query selection is providing exploration and giving chance to unlikely states to be queried at the beginning of the RBI process. In fact, for the same given task history, the proposed $f_Q$ in (\ref{eq:Renyi_monetum_RBI}) enhances the possibility of the target state selection in the query subset compared to the $f_Q$ objective in (\ref{eq:Renyi_RBI}), while the system is dealing with adversarial cases.
\newline
\begin{prop}
    \label{prop:MomentumIsBetter}
    Given $p(\ell = 1 \lvert \sigma,\phi)\in \lbrace 0,1 \rbrace$, $\forall (\sigma,\phi)$ for $\alpha \in [0,\infty)$ and $\lambda \geq 0$, given  $a,b\in\mathcal{A}$ and $r,q\in\mathcal{Q}$, where $a$ is the state of the target, $a \neq b$, and $r \neq q$, if $ \ \exists \ \mathcal{H}_{s-1} \ \text{s.t.} \ p(a\lvert\mathcal{H}_{s-1})<p(b\lvert\mathcal{H}_{s-1})$, then
    \begin{equation*}
        \label{eq:adverserial}
         \begin{array}{ll}
         p\Big(-H_{\alpha}(\sigma| \eps, \phi=q, \mathcal{H}_s) + \lambda M_{\alpha}(\phi=q,\mathcal{H}_s) \\
         \hspace{.5in} > -H_{\alpha}(\sigma| \eps, \phi=r, \mathcal{H}_s) + \lambda M_{\alpha}(\phi=r,\mathcal{H}_s)\Big)\\ [.1in]
        \geq  p \Big(-H_{\alpha}(\sigma| \eps, \phi=q, \mathcal{H}_s)  >  -H_{\alpha}(\sigma| \eps, \phi=r, \mathcal{H}_s) \Big)
        \end{array}
    \end{equation*}
\end{prop}
Proposition~\ref{prop:MomentumIsBetter} shows that the probability of $a$ (assumed to be the target state) having a higher $f_Q$ value than $b$ is larger when (\ref{eq:Renyi_monetum_query}) is used instead of $f_Q$ in (\ref{eq:Renyi_RBI}). Although, the probability of $a$ given the task history is lower than the probability of $b$ given the task history. This means that even if $a$ is less likely than $b$ according to the prior (adversarial case) and observed evidence, using the proposed query selection objective, $a$ has more chance to appear in the query subset, when compared to using \eqref{eq:action_value_mmi}. The proof of Proposition~\ref{prop:MomentumIsBetter} can be found in the supplementary material. 

It should be noted that by collecting more evidence through a recursive process, $\lambda_2$ should be dynamically updated such that the emphasis on $H_{\alpha_2}$ is increased with the number of sequences that means the $\lambda_2$ value should be decreased as the number of sequences is increasing~\cite{bilmes2000dynamic}.
\SimplexProbTrajectories
\section{Geometrical Representation}
\label{sec:Geometrical Representation}
To offer better understanding of the proposed active framework, a geometrical illustration of the RBI problem using \textit{probability simplex} is discussed in this section. Although probability simplex is widely used for Bayesian inference interpretation, to the best of our knowledge, the use of simplex to illustrate a recursive Bayesian inference through the querying process has never been studied before. By taking advantage of such a geometrical representation, we show that the estimation problem can be conceptualized as a posterior trajectory movement tracking problem, in which we gain new probabilistic insights on the decision boundary, direction and speed of the movement toward a particular state during an active inference. 

Probability mass functions can be visualized geometrically as points in a vector space, called a probability simplex, with the axes given by random variables. Figure~\ref{fig:simplex_trajectory} illustrates an example of a probability simplex, where each vertex corresponds to a state with its single probability value set to 1. For simplicity, here we visualize all the findings for fixed cardinality $|\mathcal{A}|$=3. For any given $\tau'$, stopping criterion $(g_I)$ forms a feasible set in the simplex and hence determines whether the system should keep querying or make a decision. Therefore, at any sequence $s$, if $p(\sigma | \mathcal{H}_s)$ is located in any feasible region (dashed areas), the inference is terminated and the vertex of that region is selected as the target state. 
\subsection{Momentum and Posterior Trajectory Movement}
\label{sec:Posterior Trajectory Movement}
As discussed in Section~\ref{sec:An Objective function of Posterior Changes Trajectory}, Momentum functions in $(I)$ and $(Q)$ are functions of the posterior changes. In the query optimization, Q-Momentum is a function of the vertex-directed posterior changes in the simplex, which directly influences the direction of the posterior trajectory movement in the simplex. Let's consider a simple adversarial example in Figure~\ref{fig:simplex_trajectory} where the initial probability point (provided by prior information) is spatially distant from the corner of interest. Assuming $\textbf{\textit{a}}$ is the target state, the goal is to move the posterior toward vertex $\textbf{\textit{a}}$ in the simplex. Using the Bayes rule, we can show that the posterior, at each sequence, can only move along three lines passing through vertices in the simplex space. In proposition~\ref{prop:colinearity}, we are demonstrating this geometrical fact.
\begin{prop}
\label{prop:colinearity}
For $\sigma, \phi \in \mathcal{A}=\{a_1,a_2,\dots,a_{|\mathcal{A}|}\}$, probability points $P(\phi)$, $p_{s}$, and $p_{s-1}$ are collinear, where $p_{s} = (p(a_1|\mathcal{H}_{s}),p(a_2|\mathcal{H}_{s}),\dots,p(a_{|\mathcal{A}|}|\mathcal{H}_{s}))$ and $P(\phi) = (0,0,\dots,1,\dots,0)=\mathbf{1}(\phi=\sigma)$, and $\mathbf{1}$ is an indicator operator.
\end{prop}

According to Proposition~\ref{prop:colinearity}, prior, posterior and the vertex point corresponding to the query are collinear (proof can be found in the supplementary material). 
%(\textcolor{red}{It is not clear what is the significance of the results presented in Proposition 2. Make a connection with the below paragraph and describe how this results provides a bettern understanding of the proposed active RBI framework}) 
Accordingly, $\eps_s$ and $\phi_s$ (vertex) determine the direction and movement path of $p(\sigma | \mathcal{H}_s)$ in the simplex, respectively. By employing these geometrical properties, we can demonstrate the trajectory of the posterior changes over sequences in the simplex, which provides a geometric tool to interpret and assesses the RBI process.

As discussed before, by minimizing conditional entropy $H_{\alpha}$ the querying selects the candidate with highest posterior probability (closest vertex) and fluctuates between two vertices of non-interest. Alternatively the proposed Q-Momentum measures the average change over passed sequences for each query. Going back to the adversarial example in Figure~\ref{fig:simplex_trajectory}, we can see that at beginning of the process, the system is more likely to query states $\textbf{\textit{c}}$ and $\textbf{\textit{b}}$ according to the latest posterior probability. Although the initial responses to querying both $\textbf{\textit{c}}$ and $\textbf{\textit{b}}$ are negative, we can see that the system keeps querying them until $s=4$. The proposed $f_Q$, however, has a higher probability of selecting a query aligned with vertex $\textbf{\textit{a}}$ as we stated in Proposition~\ref{prop:MomentumIsBetter}. This presumably yields a faster estimation and prevents the system from zigzagging between incorrect estimates. 
\subsection{Momentum and Posterior Movement Speed}
As another geometric insight to the RBI problem, we can see that the example presented in Figures~\ref{fig:simplex_probChanges} and \ref{fig:simplex_stopping} shows two different querying sequences, where $p(a | \mathcal{H}_s) < 1$ and $H_{\alpha} < \tau'$. Although the posterior is not placed in the feasible region, in \ref{fig:simplex_stopping}, the system is fairly close to the vertex $\textbf{\textit{a}}$. The proposed I-Momentum objective in $g_I$ provides a relaxation based on the prediction made by $D_{\alpha}$ for the next position of $p_{s+1}$ in the simplex, such that the system can make a decision even before crossing the decision boundary. This relaxation accelerates the inference process by including the history of the posterior trajectory and approximating the speed of the posterior movement. This approximation helps the system to foresee whether the next location of the posterior will be located in the feasible region or not.
%(\textcolor{red}{How exactly? Please be more specific}). 
Therefore, it is not always necessary to spend extra queries only to reach the feasible region. Detailed discussion about decision boundary can be found in the next Section. 
\section{Decision Boundary and Feasible Set in RBI}
\label{sec:Decision Boundary and Feasible Set in RBI}
Stopping criterion $g_I$ relies on the constraint of the inference step. This constraint forms a feasible set in the simplex and hence determines whether the system should continue querying or make a decision. Applying a pre-defined confidence level $\tau$ over posterior probability in $f_I$, constructs a decision boundary visualized in Figure~\ref{fig:decsionBoundary} in the probability domain. Therefore, for any sequence $s$, if $p(x) \geq \tau$ at a particular vertex, that vertex is the estimated target.
\decsionBoundary
Shaded region in Figure~\ref{fig:decsionBoundary} corresponds to the feasible set as a function of $\tau$, where $S=\{ p(X) \ \lvert \ p(x^*)\geq\tau \}$. As expected, decreasing $\tau$ enlarges the feasible set in the estimation problem. 

In the RBI problem, the query objective attempts to select a subset of queries by minimizing the conditional entropy across all directions, where the selected queries should push the posterior toward one of the triangle vertices in Figure~\ref{fig:decsionBoundary}. There exists a sequence $s$, where $H_s(X) \leq \tau''$ and $p_s(x)\geq \tau$. To investigate the relationship between $\tau''$ and $\tau$, again we will take advantage of geometrical representation in simplex. 

Figure~\ref{fig:decsionBoundary} shows $H(X)=\tau''$ and $p(a)=\tau$ boundaries. We can see that, by replacing the posterior constraint in $f_I$ with entropy function, we are in turn enlarging the feasible set. Furthermore, it illustrates that the $H_s(X)$ contour is intersecting with $p(a)=\tau$ at midline. The following Lemma proves this fact.
\begin{lemma}
\label{lemma:simplex}
For $\tau \geq \displaystyle{\frac{1}{|\mathcal{A}|}}$ and $|\mathcal{A}| > 1$, decision boundaries $p(a)=\tau$ and $H(X) \leq \tau''$ intersect when entropy is at its maximum $\tau''$, at the midpoint of line $p(a)=\tau$, $P_m$, where $p_{m_k}, \forall k \in \mathcal{A}$ is defined as follows;
\begin{equation*}
p_{m_k} =
  \begin{cases}
    \tau       & k = a\\
    \displaystyle{\frac{1-\tau}{|\mathcal{A}|-1}}  & k \neq a
  \end{cases}
\end{equation*}
and 
\begin{equation*}
\label{eq:intersection}
    \begin{array}{rl}
        \tau'' &= -\tau\log{\left(\tau\right)}-(1-\tau) \log{\left(\displaystyle{\frac{1-\tau}{|\mathcal{A}|-1}}\right)}. \\[.1in]
        \end{array}
\end{equation*}
\end{lemma}
The  proof of  Lemma~\eqref{lemma:simplex} can  be  found  in  the supplementary material. 

Subsequently, if $n$ denotes the dimensionality of the state space, we can define the intersection point as,
\begin{equation*}
    v_n(\tau)= [\tau, \displaystyle{\frac{1-\tau}{n-1}},\cdots, \displaystyle{\frac{1-\tau}{n-1}}]
\end{equation*}
Additionally, let us define the following point;
\begin{equation*}
    w_n(\tau)= [\tau, {1-\tau},0,\cdots, 0]
\end{equation*}
where this point can be generalized to the other vertices in the simplex by changing the position of $1-\tau$ with any of the $0$s. 

According to the Rényi entropy definition, we have 
\begin{equation*}
\begin{array}{cc}
H_\alpha(v_n(\tau)) \geq H_\alpha(w_n(\tau)) \\[.1in]
H_\infty(v_n(\tau)) =H_\infty(w_n(\tau))=\tau \\
\end{array}
\end{equation*}
Therefore, the decision boundary is a linear surface in the simplex. Moreover, $H_\alpha(v_n(\tau))-H_\alpha(w_n(\tau))$ is correlated with the extent of the difference between feasible regions. 
\begin{equation}
    \begin{array}{lr}
        H_\alpha(v_n(\tau))-H_\alpha(w_n(\tau)) = \\[.1in]
        \displaystyle{\frac{1}{1-\alpha}}\log \displaystyle{\frac{\tau^\alpha+(n-1)^{1-\alpha}(1-\tau)^{\alpha-1})}{\tau^\alpha + (1-\tau)^\alpha}} \\
    \end{array}
\end{equation}
Specifically for $\alpha=1$, where
\begin{equation*}
    \begin{array}{lr}
       H(v_n(\tau))-H(w_n(\tau)) = (1-p)\log(n-1).\\
    \end{array}
\end{equation*}
Figure~\ref{fig:diffRenyiEntropy} illustrates the changes of entropy as a function of $\alpha$ and the state space dimension $n$. While the inference objective $f_I$ aims to select the most likely state candidate, the decision criterion $g_I$ prevents $f_I$ from picking a high-entropy (ambiguous) state. Choosing an ambiguous candidate can be easily avoided by making sure that the decision boundary is intersecting with one of the surfaces. 

\diffRenyiEntropy
\diffRenyiAlpha
Figure~\ref{fig:diffRenyiAlpha} shows the effect of different $\alpha$ values where $n=30$ in (a) and the effect of different $n$ values where $\alpha=1$. Here, $\tilde{\tau}$ is defined such that for a given $\alpha$, $H_{\alpha}(v_n(\tau))=H_{\alpha}(w_n(\tilde{\tau}))$, where decision boundary intersects with the surface. It should be noted that we cannot always find a $\tilde{\tau}$ value such that for a given $\tau$, the equi-entropy contours completely contained within the simplex.
%\textcolor{red}{It is kind of confusing here. Are we using the posterior distribution or the conditional entropy as $f_I$ ? We need to clearly write this section and specify what we use as $f_I$ and why... } 
It can be observed from both plots in Figure~\ref{fig:diffRenyiAlpha} that for the selected values for $\tilde{\tau}$,  equi-entropy balls reside inside the simplex and can force a termination in the estimation not getting closer to one vertex. Therefore $\alpha$ can be chosen accordingly with $\tau$ value and the cardinality of the state space $n$. For instance, in the case of a high confidence level, e.g. $\tau \approx 0.8$ with $n=30$, we should avoid selecting an  $\alpha$ such that  $\alpha << 1$, to prevent dramatic performance loss.
%
%Here, we introduce a key value $\tilde{\tau}$ for $w_n(.)$ to represent a point where decision boundary intersects with the surface. Accordingly $\alpha$, it is possible to have a $\tilde{\tau} \ \text{such that} \ H_{\alpha}(v_n(\tau))=H_{\alpha}(w_n(\tilde{\tau}))$. Note that it is also possible not to find a $\tilde{\tau}$ value indicating the equi-entropy contours for the value $\tau$ completely contained within the simplex.
%It should be noted that for a given $\alpha$, there might be a $\tilde{\tau} \ \text{such that} \ H_{\alpha}(v_n(\tau))=H_{\alpha}(w_n(\tilde{\tau}))$. It is apparent equi-entropy balls can yield completely inside the simplex and hence does not intersect with any of the surfaces.
%
\section{Experimental Results}
\label{sec:Experimental Results}
% Results
To evaluate the performance of the proposed active inference framework, first we study role of $\alpha$ and $\lambda$ parameters in the recursive inference process and show how using the unified framework can jointly enhance both speed and precision in a RBI problem. Here, the accuracy and speed are defined as the number of correct selections and the inverse number of queries required for the inference problem. In addition to that, we compare the performance of the proposed framework to the other approaches discussed in the introduction such as MMI and random selection in two applications. 
\subsection{Speed-accuracy Trade-off}
\label{sec:Impact of alpha}
At a more general level, for the first empirical analysis we show some simulation results regarding the impact of $\alpha$ in the inference performance. For this study, we employed a simple target detection task using Monte-Carlo sampling. The goal of the task is to guess the target state (out of 30 candidates) by querying and collecting samples from two class conditional distributions, i.e. target and non-target classes. To simplify the sampling process, we assume evidence is sampled from Gaussian distributions conditioned on query and state tuples. Specifically we select these distributions to be 1D and these distributions are further used to calculate the posterior distribution over the states after each recursion.

\simulationAlpha
\speedAccQuery
\simplexSimulations
To evaluate the evolution of the posterior under different query selection methods, we have compared the proposed method with five proposed active query selection methods in~\cite{tong2001support,dasgupta2008hierarchical,wilson2012bayesian,sourati2017probabilistic,higger2017recursive} as follows. First, we implemented the proposed active query selection proposed by Tong and Koller in~\cite{tong2001support} that aims to maximize the state-space spanned with a pre-defined kernel at a given location. For our classification problem, it is assumed that the state space is discrete with dirac-delta function as the kernel. Then we have had the Fisher information-based method proposed by Sourati in~\cite{sourati2017probabilistic} and the MMI-based method proposed by Higger in~\cite{higger2017recursive}. We also implemented the expected posterior maximization method proposed by Wilson in~\cite{wilson2012bayesian}, in which it is assumed that the class conditional evidence distributions are known. Finally, we have implemented the hierarchical sampling method proposed by Dasgupta and Hsu in~\cite{dasgupta2008hierarchical}. Here, to balance the exploration and exploitation process, a percentage of queries are selected to minimize uncertainty and the rest are selected randomly.
%
%In the work by Tong \cite{tong2001support}, propose an active query selection paradigm which aims to maximize the state-space spanned with a predefined kernel at a given location. In classification problem it is convenient to assume the stat space is discrete with dirac-delta function as the kernel. Sourati in the work \cite{sourati2017probabilistic}, promotes Fisher information ratio to select a subset of training samples. With the use of Fisher information selected samples, the author aims to select samples that shift the parameters towards the most informative point. Trivially, in a classification task, the parameter is the argument of the maximum posterior and hence this idea can be adopted to maximize the information of the argument in the posterior. Additionally hierarchical sampling is also used in active learning, specifically, one forms a query subset partitioned with different objectives. For example, to balance the exploration and exploitation, a percentage of queries are selected to minimize uncertainty and rest are selected randomly \cite{dasgupta2008hierarchical}. Wilson in the work \cite{wilson2012bayesian}, proposes sampling with a Monte Carlo estimate dependent variational distance measure. In our work we already assume the class conditional evidence distributions are known through the inference and hence the method is equivalent to minimizing conditional entropy. Higger in the work \cite{higger2017recursive}, elaborates usage of mutual information based query selection.

Figure~\ref{fig:simulation_diff_alpha} illustrates the target state posterior evolution in estimation over sequences under three different conditions on the prior probability $(\mathcal{H}_0)$ at the beginning of the process including uniform, adversarial, and supportive priors. Taking advantages of the history of the posterior changes, in the adversarial case, the proposed method outperforms the other methods in terms of both speed and accuracy. For the other two cases (non-adversarial), we can observe similar performance to the MMI method. As another observation, we can see that depending on the prior probability, the inference process can be influenced differently by using different values of $\alpha_2 < 1$ and $\alpha_2 > 1$. As a justification, consider the Q-Momentum term in (\ref{eq:Renyi_monetum_query}). If $\alpha_2 < 1$, maximizing $\displaystyle{\frac{1}{(\alpha_2 - 1)}}\log{\displaystyle{\sum_{\sigma \in \mathcal{A}}}\displaystyle{\frac{p^{\alpha_2}(\sigma|\eps, \phi, \mathcal{H}_{n})}{p^{{\alpha_2}-1}(\sigma|\mathcal{H}_{n})}}}$ is equivalent to minimizing $\displaystyle{\frac{1}{(1 - \alpha_2)}}\log{\displaystyle{\sum_{\sigma \in \mathcal{A}}}}{p^{\alpha_2}(\sigma|\eps, \phi, \mathcal{H}_{n})p^{{1-\alpha_2}}(\sigma|\mathcal{H}_{n})}$. Accordingly, the system will query a subset of most unlikely states based on the $\mathcal{H}_0$, which helps the posterior to moves toward the unlikely target state (state 16) faster than $\alpha_2 \geq 1$ and MMI.
Using the similar simulation study, Figure~\ref{fig:speedAccQuery} shows the speed-accuracy changes as a function of $\alpha_{1,2}$ and $\lambda_{1,2}$ parameters in (\ref{eq:Renyi_monetum_RBI}).

Figures~\ref{fig:stopping} and \ref{fig:querying} illustrate speed-accuracy changes as a function of $(\alpha_{1}, \lambda_{1})$ and $(\alpha_{2}, \lambda_{2})$, respectively. Here, the size of markers is a linear function of accuracy and speed and larger points shows higher speed and accuracy. Comparing the speed and accuracy of points obtained from unified method to the conventional case, i.e. $\alpha_2=1$, $\lambda_2=0$, we can see that adjusting these parameters in RBI allows us to increase speed and accuracy simultaneously.

Figure~\ref{fig:simplexSimulations} demonstrates the generality of the discussed geometric approach in Section~\ref{sec:Geometrical Representation} and connects the discussed geometrical representation with the experimental part. Figure~\ref{fig:simplexquery} illustrates recursive posterior transition in the simplex using different query selection methods. The same experimental settings used for speed-accuracy analyses is used in generating ~\ref{fig:simplexquery}, except for estimating the target among 3 candidates as opposed to 30. The analysis uses two 1-dimensional Gaussian distributions, $\mathcal{N}(0, 1)$, $\mathcal{N}(3, 1.5)$ to model target and non-target evidence distributions. Moreover, Figure~\ref{fig:SimplexSpeed} shows the impact of the prior probability on the posterior movement trajectories by showing the average trajectory and its variability for 500 Monte-Carlo simulations in the simplex.
\subsection{Restaurant Recommender}
\label{sec:Restaurant Recommender}
The primary aim in all recommender systems is to understand what the customer is looking for. For restaurant recommendation, assuming that the diner has a certain type of restaurant in mind, the system needs to discover their preferred restaurant according to the diner's profile and their responses to the asked queries. Given the projection matrix for mapping the queries to the state (extracted from query understanding), we can use the proposed inference method to estimate the diner's intent. Noted that extracting the projection matrix via query understanding is outside the scope of this paper. Here, we used Entree Chicago Recommendation dataset~\cite{Dua:2019}, which consists of list of restaurant, cuisines, users' scores. Using the dataset, we learned diners' profiles and projection matrix between list of queries and places. By mapping diners' scores to binary \textit{Yes} and \textit{No} answers, we used Monte-Carlo simulation to draw samples from two conditional distributions over 80 restaurants. Figure~\ref{fig:food_recommender} shows an example of the performance of the recommender system for two diners using the proposed method, MMI and random approaches. Figure~\ref{fig:food_supportive} belongs to a diner with supportive profile, which means this diner prefers similar places that they have tried before (a conventional diner). Figure~\ref{fig:food_adversarial} depicts an adversarial profile where the diner is eager to try new places (an adventurous diner). Applying the proposed method over 50 diners shows that on average we improve accuracy and speed by \textbf{8\%} and \textbf{10\%} respectively in the user intent estimation. We present our results in Table~\ref{table:resultsTable}. In all of the studies, we decrease $\lambda_2$ with a small annealing rate over sequences down to a minimum value $0$.
%
%\simulationSpeedAcc
%
\foodRecommender
\resultsTable
\subsection{Brain-computer Interface (BCI) Typing System}
As another application, we use a language-model-assisted EEG-based BCI typing interface~\cite{akcakaya2013noninvasive}. User intent detection is one of the main components of a BCI system, in which the system estimates the intended user state according to the brain activities, e.g. electroencephalogram (EEG) evidence collected under specified presentation mode and prior knowledge provided by the language model (LM). In order to estimate the posterior probability over typing symbols such as English alphabet, a context prior $P\left(\sigma|\mathcal{H}_0\right)$ is provided by the LM, which estimates the conditional probability of every letter in the alphabet based on $n-1$ previously typed letters in a Markov model framework. The BCI-based typing interfaces typically use a rapid serial visual presentation, in which the system uses a finite set of symbols. In this study, both $\sigma$ and $\phi$ belong to $\mathcal{A} = \mathcal{Q} = \{A, B, C, ..., Z, \_, < \}$, where $\_$ and $<$ represent space and backspace, respectively. As observation, EEG signals were acquired from 10 sensors according to International 10-20 System locations: Fp1, Fp2, Fz, F3, F4, F7, F8, Cz, C3, C4, T3, T4, T5, T6, P3, P4, O1, O2, A1 and A2. A DSI-24 Wearable Sensing EEG Headset was used for data acquisition, at a sampling rate of 300 Hz with active dry electrodes. All participants performed the \textit{calibration} session containing 100 sequences; each sequence includes 8 trials (queries); and one trial in each sequence is the target symbol which is displayed on the screen prior to each sequence. The time interval between trials is 200 ms. Optimal parameters for both target and non-target class distributions were learned using the calibration data, which are used in simulation studies and \textit{copy phrase} task.

BCI typing systems require real-time stimuli sequence optimization. Query optimization for BCI typing systems is not a well-studied problem. To the best of our knowledge, there are limited number of studies that addressed the query optimization problem for the BCI typing system designs~\cite{higger2017recursive, mainsahBCI2018} and all of the proposed query selection methods are using MMI method that results in the selection of the \textit{N-best} symbols based on the posterior distribution conditioned on the LM~\cite{koccanaogullari2018analysis}. There are cases that posterior conditioned on the LM and EEG evidence are misleading especially if the intended symbol is highly unlikely for the LM. We show that using the proposed unified framework enhances the typing performance measures in compare to MMI method, which only exploits the current knowledge based on the posterior.

\textbf{Simulation Study:}
To evaluate the empirical performance of the proposed active inference, a copy phrase task was simulated using EEG data collected during the calibration sessions. Using class conditional distributions $f_{\sigma^*, \phi^i_s}$ (target class) and $f_{\not\sigma^*, \phi^i_s}$ (non-target class), we used Monte-Carlo sampling method to draw samples from class distributions for typing letter. For instance, the target phrase is \textit{"She needs one month to \textbf{convalesce}."} and \textbf{convalesce} is the target word which needs to be typed to complete the phrase.

Figure~\ref{fig:bciTyping} shows the average of two standard typing accuracy and speed measures such as accuracy in typing a letter correctly (ATL) or information transfer rate (ITR) for 10 healthy participants. In this task, Proposed-v$\alpha$ again provided the highest speed and accuracy.
The implemented BCI system is publicly accessible from \url{https://github.com/BciPy/BciPy}.
%Detailed information for this application and performance measurement part can be found in supplementary material. 
%
%For our simulations we use the system presented in \cite{memmott2020bcipy}
%
\BCItyping
\section{Conclusions}
A new active inference framework unifying the query selection and decision making objectives of recursive Bayesian inference has been proposed. More specifically, being motivated by information theoretic approaches, we have unified the RBI with a new objective based on the $\alpha$-entropy and $\alpha$-divergence. We have provided a geometrical interpretation of the RBI over the probability simplex to demonstrate the progression of belief over the states during inference and query selection steps. We showed both theoretically and through three different testbeds that the proposed framework improves both speed and accuracy of state estimation.
%
%\section{Acknowledgment}
%This work is supported by NSF (IIS-1149570, CNS-1544895, IIS-1715858, IIS-1717654, IIS-1844885), DHHS (90RE5017-02-01), and NIH (R01DC009834).
%
\bibliographystyle{IEEEtran}
%\scriptsize
\bibliography{references}

% Generated by IEEEtran.bst, version: 1.14 (2015/08/26)
\begin{thebibliography}{10}
\providecommand{\url}[1]{#1}
\csname url@samestyle\endcsname
\providecommand{\newblock}{\relax}
\providecommand{\bibinfo}[2]{#2}
\providecommand{\BIBentrySTDinterwordspacing}{\spaceskip=0pt\relax}
\providecommand{\BIBentryALTinterwordstretchfactor}{4}
\providecommand{\BIBentryALTinterwordspacing}{\spaceskip=\fontdimen2\font plus
\BIBentryALTinterwordstretchfactor\fontdimen3\font minus
  \fontdimen4\font\relax}
\providecommand{\BIBforeignlanguage}[2]{{%
\expandafter\ifx\csname l@#1\endcsname\relax
\typeout{** WARNING: IEEEtran.bst: No hyphenation pattern has been}%
\typeout{** loaded for the language `#1'. Using the pattern for}%
\typeout{** the default language instead.}%
\else
\language=\csname l@#1\endcsname
\fi
#2}}
\providecommand{\BIBdecl}{\relax}
\BIBdecl

\bibitem{bergman1999recursive}
N.~Bergman, ``Recursive bayesian estimation,'' \emph{Department of Electrical
  Engineering, Link{\"o}ping University, Link{\"o}ping Studies in Science and
  Technology. Doctoral dissertation}, vol. 579, p.~11, 1999.

\bibitem{sarkka2013bayesian}
S.~S{\"a}rkk{\"a}, \emph{Bayesian filtering and smoothing}.\hskip 1em plus
  0.5em minus 0.4em\relax Cambridge University Press, 2013, vol.~3.

\bibitem{wilson2012bayesian}
A.~Wilson, A.~Fern, and P.~Tadepalli, ``A bayesian approach for policy learning
  from trajectory preference queries,'' in \emph{Advances in neural information
  processing systems}, 2012, pp. 1133--1141.

\bibitem{hoi2006batch}
S.~C. Hoi, R.~Jin, J.~Zhu, and M.~R. Lyu, ``Batch mode active learning and its
  application to medical image classification,'' in \emph{Proceedings of the
  23rd international conference on Machine learning}.\hskip 1em plus 0.5em
  minus 0.4em\relax ACM, 2006, pp. 417--424.

\bibitem{chepuri2015sparsity}
S.~P. Chepuri and G.~Leus, ``Sparsity-promoting sensor selection for non-linear
  measurement models,'' \emph{IEEE Transactions on Signal Processing}, vol.~63,
  no.~3, pp. 684--698, 2015.

\bibitem{sourati2017probabilistic}
J.~Sourati, M.~Akcakaya, D.~Erdogmus, T.~K. Leen, and J.~G. Dy, ``A
  probabilistic active learning algorithm based on fisher information ratio,''
  \emph{IEEE transactions on pattern analysis and machine intelligence},
  vol.~40, no.~8, pp. 2023--2029, 2017.

\bibitem{golovin2010near}
D.~Golovin, A.~Krause, and D.~Ray, ``Near-optimal {B}ayesian active learning
  with noisy observations,'' in \emph{Advances in Neural Information Processing
  Systems}, 2010, pp. 766--774.

\bibitem{higger2017recursive}
M.~Higger, F.~Quivira, M.~Akcakaya, M.~Moghadamfalahi, H.~Nezamfar, M.~Cetin,
  and D.~Erdogmus, ``Recursive bayesian coding for bcis,'' \emph{IEEE
  Transactions on Neural Systems and Rehabilitation Engineering}, vol.~25,
  no.~6, pp. 704--714, 2017.

\bibitem{jedynak2012twenty}
B.~Jedynak, P.~I. Frazier, and R.~Sznitman, ``Twenty questions with noise:
  {B}ayes optimal policies for entropy loss,'' \emph{Journal of Applied
  Probability}, vol.~49, no.~1, pp. 114--136, 2012.

\bibitem{MoghadamJSTS}
M.~{Moghadamfalahi}, M.~{Akcakaya}, H.~{Nezamfar}, J.~{Sourati}, and
  D.~{Erdogmus}, ``{An Active RBSE Framework to Generate Optimal Stimulus
  Sequences in a BCI for Spelling},'' \emph{ArXiv e-prints (arXiv:1607.03578
  [cs.HC])}, Jul. 2016.

\bibitem{mainsahBCI2018}
B.~Mainsah, D.~Kalika, L.~Collins, S.~Liu, and C.~Throckmorton,
  ``Information-based adaptive stimulus selection to optimize communication
  efficiency in brain-computer interfaces,'' in \emph{Advances in Neural
  Information Processing Systems}, 2018, pp. 4820--4830.

\bibitem{koccanaogullari2018analysis}
A.~Ko{\c{c}}anaogullar{\i}, M.~Ak{\c{c}}akaya, and D.~Erdogmus, ``On analysis
  of active querying for recursive state estimation,'' \emph{IEEE Signal
  Processing Letters}, vol.~25, no.~6, p. 743, 2018.

\bibitem{tsiligkaridis2014collaborative}
T.~Tsiligkaridis, B.~M. Sadler, and A.~O. Hero, ``Collaborative 20 questions
  for target localization,'' \emph{IEEE Transactions on Information Theory},
  vol.~60, no.~4, pp. 2233--2252, 2014.

\bibitem{dasgupta2008hierarchical}
S.~Dasgupta and D.~Hsu, ``Hierarchical sampling for active learning,'' in
  \emph{Proceedings of the 25th international conference on Machine learning},
  2008, pp. 208--215.

\bibitem{baram2004online}
Y.~Baram, R.~E. Yaniv, and K.~Luz, ``Online choice of active learning
  algorithms,'' \emph{Journal of Machine Learning Research}, vol.~5, no. Mar,
  pp. 255--291, 2004.

\bibitem{liu2013query}
J.~Liu, P.~Pasupat, Y.~Wang, S.~Cyphers, and J.~Glass, ``Query understanding
  enhanced by hierarchical parsing structures,'' in \emph{2013 IEEE Workshop on
  Automatic Speech Recognition and Understanding}.\hskip 1em plus 0.5em minus
  0.4em\relax IEEE, 2013, pp. 72--77.

\bibitem{barron1986entropy}
A.~R. Barron \emph{et~al.}, ``Entropy and the central limit theorem.''
  \emph{Ann. Prob.}, vol.~14, no.~1, pp. 336--342, 1986.

\bibitem{johnson2004fisher}
O.~Johnson and A.~Barron, ``Fisher information inequalities and the central
  limit theorem,'' \emph{Probability Theory and Related Fields}, vol. 129,
  no.~3, pp. 391--409, 2004.

\bibitem{farias2011irrevocable}
V.~F. Farias and R.~Madan, ``The irrevocable multiarmed bandit problem,''
  \emph{Operations Research}, vol.~59, no.~2, pp. 383--399, 2011.

\bibitem{renyi1961measures}
A.~R{\'e}nyi \emph{et~al.}, ``On measures of entropy and information,'' in
  \emph{Proceedings of the Fourth Berkeley Symposium on Mathematical Statistics
  and Probability, Volume 1: Contributions to the Theory of Statistics}.\hskip
  1em plus 0.5em minus 0.4em\relax The Regents of the University of California,
  1961.

\bibitem{song2001renyi}
K.-S. Song, ``R{\'e}nyi information, loglikelihood and an intrinsic
  distribution measure,'' \emph{Journal of Statistical Planning and Inference},
  vol.~93, no. 1-2, pp. 51--69, 2001.

\bibitem{li2016renyi}
Y.~Li and R.~E. Turner, ``R{\'e}nyi divergence variational inference,'' in
  \emph{Advances in Neural Information Processing Systems}, 2016, pp.
  1073--1081.

\bibitem{madiman2008entropy}
M.~Madiman, ``On the entropy of sums,'' in \emph{2008 IEEE Information Theory
  Workshop}.\hskip 1em plus 0.5em minus 0.4em\relax IEEE, 2008, pp. 303--307.

\bibitem{van2010renyi}
T.~van Erven and P.~Harremo{\"e}s, ``R{\'e}nyi divergence and majorization,''
  in \emph{2010 IEEE International Symposium on Information Theory}.\hskip 1em
  plus 0.5em minus 0.4em\relax IEEE, 2010, pp. 1335--1339.

\bibitem{koccanaougullari2018optimal}
A.~Ko{\c{c}}anao{\u{g}}ullar{\i}, Y.~M. Marghi, M.~Ak{\c{c}}akaya, and
  D.~Erdo{\u{g}}mu{\c{s}}, ``Optimal query selection using multi-armed
  bandits,'' \emph{IEEE Signal Processing Letters}, vol.~25, no.~12, pp.
  1870--1874, 2018.

\bibitem{marghi2019history}
Y.~M. Marghi, A.~Ko{\c{c}}anao{\u{g}}ullar{\i}, M.~Ak{\c{c}}akaya, and
  D.~Erdo{\u{g}}mu{\c{s}}, ``A history-based stopping criterion in recursive
  bayesian state estimation,'' in \emph{ICASSP 2019-2019 IEEE International
  Conference on Acoustics, Speech and Signal Processing (ICASSP)}.\hskip 1em
  plus 0.5em minus 0.4em\relax IEEE, 2019, pp. 3362--3366.

\bibitem{bilmes2000dynamic}
J.~A. Bilmes, ``Dynamic bayesian multinets,'' in \emph{Proceedings of the
  Sixteenth conference on Uncertainty in artificial intelligence}.\hskip 1em
  plus 0.5em minus 0.4em\relax Morgan Kaufmann Publishers Inc., 2000, pp.
  38--45.

\bibitem{tong2001support}
S.~Tong and D.~Koller, ``Support vector machine active learning with
  applications to text classification,'' \emph{Journal of machine learning
  research}, vol.~2, no. Nov, pp. 45--66, 2001.

\bibitem{Dua:2019}
\BIBentryALTinterwordspacing
D.~Dua and C.~Graff, ``{UCI} machine learning repository,'' 2017. [Online].
  Available: \url{http://archive.ics.uci.edu/ml}
\BIBentrySTDinterwordspacing

\bibitem{akcakaya2013noninvasive}
M.~Akcakaya, B.~Peters, M.~Moghadamfalahi, A.~R. Mooney, U.~Orhan, B.~Oken,
  D.~Erdogmus, and M.~Fried-Oken, ``Noninvasive brain--computer interfaces for
  augmentative and alternative communication,'' \emph{IEEE reviews in
  biomedical engineering}, vol.~7, pp. 31--49, 2013.

\end{thebibliography}
\newpage
\title{Supplementary Materials}
\section*{Supplementary Materials}
\subsection{Proof of Proposition 1}
% policy evaluation
% \begin{prop}
\textbf{Proposition  1.}
    \textit{Given, $p(\ell = 1 \lvert \sigma,\phi)\in \lbrace 0,1 \rbrace$, $\forall (\sigma,\phi)$ for $\alpha \in [0,\infty)$ and $\lambda \geq 0$, given  $a,b\in\mathcal{A}$ and $r,q\in\mathcal{Q}$, where $a$ is the state of the target, $a \neq b$, and $r \neq q$, if $ \ \exists \ \mathcal{H}_{s-1} \ \text{s.t.} \ p(a\lvert\mathcal{H}_{s-1})<p(b\lvert\mathcal{H}_{s-1})$, then}
    \begin{equation*}
        \label{eq:adverserial}
         \begin{array}{ll}
         p\Big(-H_{\alpha}(\sigma| \eps, \phi=q, \mathcal{H}_s) + \lambda M_{\alpha}(\phi=q,\mathcal{H}_s) \\
         \hspace{.5in} > -H_{\alpha}(\sigma| \eps, \phi=r, \mathcal{H}_s) + \lambda M_{\alpha}(\phi=r,\mathcal{H}_s)\Big)\\ [.1in]
        \geq  p \Big(-H_{\alpha}(\sigma| \eps, \phi=q, \mathcal{H}_s)  >  -H_{\alpha}(\sigma| \eps, \phi=r, \mathcal{H}_s) \Big)
        \end{array}
    \end{equation*}
% \end{prop}
%
\begin{proof}
Given $a,b$ as two possible states and $q,r$ as two possible queries such that $p(\ell= 1| \sigma=a,\phi=q) = p(\ell=1 | \sigma=b,\phi=r) = 1$ and $p(\ell= 1| \sigma \neq a,\phi=q) = p(\ell=1 | \sigma \neq b,\phi=r) = 0$. Observe that, according to one to one correspondence in state query tuples; $\sum_\sigma f(\sigma,q) p(\ell|\sigma,q)=f(a,q) + \sum_{\sigma\neq a} f(\sigma,q)$ where $f(\cdot,\cdot)$ represents an arbitrary function with arguments from state space and query space correspondingly. Therefore, objectives for query selection can be represented only by corresponding state variables. For the sake of simplicity we denote:
\begin{itemize}
    \item $H_{\alpha}(q) = H_{\alpha}(\sigma=a| \phi=q ,\mathcal{H}_{s})$, \\ $H_{\alpha}(r) = H_{\alpha}(\sigma=b| \phi=r ,\mathcal{H}_{s})$ 
    \item $M_{\alpha}(q) = M_{\alpha}(\phi=q , \ \mathcal{H}_{s})$, \\ $M_{\alpha}(r) = M_{\alpha}(\phi=r , \mathcal{H}_{s})$
    \item $p(a) = p(a\lvert\mathcal{H}_{s})$, \ $p(b) = p(b\lvert\mathcal{H}_{s})$
    \item $p(a|\varepsilon, q) = p(a|\varepsilon, \phi=q, \mathcal{H}_{s})$, \\ $p(b|\varepsilon, r) = p(b|\varepsilon, \phi=r, \mathcal{H}_{s})$
\end{itemize}
We can define notations for $b$ accordingly and ignore $\lambda$ as it holds $\forall \lambda \geq 0$. 
Using the probabilistic identity of sum of random variables, we have:
    \begin{equation*}
    \begin{array}{cc}
     p(A+B>C+D)\geq p(A>C,B>D)= p(A>C\vert B>D ) \\
      p(A>C\vert B>D ) = p(B>D) \ \ \forall A,B,C,D
      \end{array}
    \end{equation*}
Thus we can write;
\begin{equation*}
\begin{array}{rl}
    &p\left(-H_{\alpha}(q) + \lambda M_{\alpha}(q) \ > \ -H_{\alpha}(r) + \lambda M_{\alpha}(r) \right) \\
    &\geq p\left(M_{\alpha}(q) > M_{\alpha}(r) |-H_{\alpha}(q)>-H_{\alpha}(r)\right) \times \\  &  \ \ \ p\left(-H_{\alpha}(q)>-H_{\alpha}(r)\ \right)
    \end{array}
\end{equation*}
\noindent
Since we are dealing with an adversarial case,  $p(a|\mathcal{H}_{s})<p(b|\mathcal{H}_{s})$. Given condition $-H_\alpha(q)>-H_\alpha(r)$ and $p(a|\mathcal{H}_{s})<p(b|\mathcal{H}_{s})$, for $\alpha > 1$, we have:
\begin{equation*}
        \begin{split}
        -H_\alpha(q) &> -H_\alpha(r) 
        \end{split}
    \end{equation*}
where $-H_\alpha(\sigma|\mathcal{H}_s) = \displaystyle{\frac{1}{\alpha - 1}}\log{\left(\displaystyle{\sum_{\sigma \in \mathcal{A}} \ p^{\alpha}(\sigma|\mathcal{H}_s)} \right)}$. Accordingly,
\begin{multline*}
        \displaystyle{\frac{1}{\alpha-1}}\mathbf{E}\left[\log{\left(\displaystyle{\sum_{\sigma \in \mathcal{A}} \ p^{\alpha}(\sigma|\eps, q)} \right)}\right] > \\ \displaystyle{\frac{1}{\alpha-1}}\mathbf{E}\left[\log{\left(\displaystyle{\sum_{\sigma \in \mathcal{A}} \ p^{\alpha}(\sigma|\eps, r)} \right)}\right] \\[0.2in]
\end{multline*}
\begin{multline*}
        \displaystyle{\sum_{\sigma \in \mathcal{A}} \ p^{\alpha}(\sigma|\eps, q)} > \displaystyle{\sum_{\sigma \in \mathcal{A}} \ p^{\alpha}(\sigma|\eps, r)}\\
\end{multline*}
Since for $\alpha > 1$, $p^{\alpha-1}(a) < p^{\alpha-1}(b)$, therefore,
\begin{equation*}
        \begin{split}
        \displaystyle{\frac{\displaystyle{\sum_{\sigma \in \mathcal{A}} \ p^{\alpha}(\sigma|\eps, q)}}{p^{\alpha-1}(a)}} &> \displaystyle{\frac{\displaystyle{\sum_{\sigma \in \mathcal{A}} \ p^{\alpha}(\sigma|\eps, r)}}{p^{\alpha-1}(b)}}\\[.1in]
        \displaystyle{\sum_{\sigma \in \mathcal{A}}} \displaystyle{\frac{\ p^{\alpha}(\sigma|\eps, q)}{p^{\alpha-1}(a)}}&> \displaystyle{\sum_{\sigma \in \mathcal{A}}} \displaystyle{\frac{\ p^{\alpha}(\sigma|\eps, r)}{p^{\alpha-1}(b)}}\\[.1in]
        \displaystyle{\frac{1}{\alpha -1}}\log \displaystyle{\sum_{\sigma \in \mathcal{A}}} \displaystyle{\frac{\ p^{\alpha}(\sigma|\eps, q)}{p^{\alpha-1}(a)}}&>  \displaystyle{\frac{1}{\alpha -1}}\log \displaystyle{\sum_{\sigma \in \mathcal{A}}} \displaystyle{\frac{\ p^{\alpha}(\sigma|\eps, r)}{p^{\alpha-1}(b)}}\\
        \end{split}
    \end{equation*}
Since, $p(\ell= 1| \sigma \neq a,\phi=q) = p(\ell=1 | \sigma \neq b,\phi=r) = 0$, according to the definition of $M_{\alpha}$, we can conclude $M_\alpha(q) > M_\alpha(r)$. Since for $0<\alpha<1$, $p^{\alpha-1}(a) > p^{\alpha-1}(b)$, the same conclusion can be drawn.    

Accordingly, when $p(a) < p(b)$ and $-H_\alpha(q) > -H_\alpha(r)$, we have $M_\alpha(q) > M_\alpha(r)$, which means $p\left(M_{\alpha}(q) > M_{\alpha}(r) \ | \ -H_{\alpha}(q)>-H_{\alpha}(r) \right)=1$ for the adversarial case, and
\begin{equation*}
\begin{array}{rl}
    &p\left(-H_{\alpha}(q) + \lambda M_{\alpha}(q) \ > \ -H_{\alpha}(r) + \lambda M_{\alpha}(r) \right) \\[.1in]
    &\hspace{.5in}\geq p\left(-H_{\alpha}(q) \ > \ -H_{\alpha}(r)\right)
    \end{array}
\end{equation*}
\end{proof}
\noindent
\subsection{Proof of Proposition 2}
\textbf{Proposition 2.}
\textit{for $\sigma, \phi \in \mathcal{A}=\{a_1,a_2,\dots,a_{|\mathcal{A}|}\}$, probability points $P(\phi)$, $p_{s}$, and $p_{s-1}$ are collinear, where $p_{s} = (p(a_1|\mathcal{H}_{s}),p(a_2|\mathcal{H}_{s}),\dots,p(a_{|\mathcal{A}|}|\mathcal{H}_{s}))$ and $P(\phi) = (0,0,\dots,1,\dots,0)=\mathbf{1}(\phi=\sigma)$, and $\mathbf{1}$ is an indicator operator.}
\begin{proof}
Using Bayes' rule, we have,
\begin{equation*}
\begin{array}{rl}
\label{eq:bayess1}
    p(\sigma=a_i|\eps, \phi=a_j,\mathcal{H}_{s-1}) &=  p(\sigma|\mathcal{H}_{s-1})\displaystyle{\frac{p(\eps|a_i,a_j)}{p(\eps|a_j)}} \\[.1in]
\end{array}
\end{equation*}
\begin{equation*}
\begin{array}{rl}
\label{eq:bayess2}
    p(\sigma=a_i|\eps, \phi=a_j,\mathcal{H}_{s-1})&=p(\sigma|\mathcal{H}_{s-1})\displaystyle{\frac{p(\eps|l=0)}{p(\eps|a_j)}} , \\ & \ \ \forall i,j \in [1,N], i \neq j\\[.1in]
\end{array}
\end{equation*}
\begin{equation*}
\begin{array}{rl}
\label{eq:bayess3}
    p(\sigma=a_i|\eps, \phi=a_i,\mathcal{H}_{s-1})&=p(\sigma|\mathcal{H}_{s-1})\displaystyle{\frac{p(\eps|l=1)}{p(\eps|a_i)}}, \\ &  \ \ \forall i \in [1,N] \\[.1in]
\end{array}
\end{equation*}
Using the fact that all probability points located in the probability simplex have the property $\sum_{i}p(a_i|\mathcal{H}_s)=1$, 
\begin{equation}
\begin{array}{rl}
    k_1=\displaystyle{\frac{p(\eps|l=1)}{p(\eps|\phi=a_j)}}, \ k_2=\displaystyle{\frac{p(\eps|l=0)}{p(\eps|\phi=a_j)}} 
\end{array}
\end{equation}
\begin{equation}
\begin{array}{rl}
\label{eq:k1k2}
    k_2=\displaystyle{\frac{1-p(\phi=a_j|\mathcal{H}_{s-1})k_1}{1-p(\phi=a_j|\mathcal{H}_{s-1})}}\\[.1in]
\end{array}
\end{equation}
In geometry, any two points $X \in \mathcal{R}^N$ $X_1=(x_{1_1},x_{1_2},\dots,x_{1_N})$, $X_2=(x_{2_1},x_{2_2},\dots,x_{2_N})$ are trivially collinear. The line containing these two points is defined as:
\begin{equation}
\begin{split}
    \frac{x_1 - x_{1_1}}{x_{2_1}-x_{1_1}}= \frac{x_2 - x_{1_2}}{x_{2_2}-x_{1_2}} = \dots = \frac{x_N - x_{1_N}}{x_{2_N}-x_{1_N}}
\end{split}
\end{equation}
Accordingly, for $|\mathcal{A}|=N$ the line determined by $p_0$ and $p_1$ is defined as:
\begin{equation}
\begin{split}
 (l_1): \ \ &\frac{x_1 - p(a_1|\mathcal{H}_{s-1})}{p(a_1|\mathcal{H}_{s})-p(a_1|\mathcal{H}_{s-1})}= \dots= \\
 &\frac{x_j - p(a_j|\mathcal{H}_{s-1})}{p(a_j|\mathcal{H}_{s})-p(a_j|\mathcal{H}_{s-1})}= \dots = \\
 &\frac{x_N - p(a_N,|\mathcal{H}_{s-1})}{p(a_N,|\mathcal{H}_{s})-p(a_N,|\mathcal{H}_{s-1}))}
\end{split}
\end{equation}
According to the relationship between $p(a_i|\mathcal{H}_{s-1})$ and $p(a_i|\mathcal{H}_{s})$ in (\ref{eq:bayess2}) and (\ref{eq:bayess3}),
\begin{equation}
\begin{split}
 &\frac{x_1 - p(a_1|\mathcal{H}_{s-1})}{p(a_1|\mathcal{H}_{s-1})(k_2-1)}= \dots= \\
 &\frac{x_j - p(a_j|\mathcal{H}_{s-1})}{p(a_j|\mathcal{H}_{s-1})(k_1-1)}= \dots = \\
 &\frac{x_N - p(a_N|\mathcal{H}_{s-1})}{p(a_N|\mathcal{H}_{s-1})(k_2-1)}
\end{split}
\end{equation}
If $x=P(\phi=a_j)$ point lies on $(\l_1)$ line, then
\begin{equation}
\begin{split}
    &\frac{0 - p(a_1|\mathcal{H}_{s-1})}{p(a_1|\mathcal{H}_{s-1})(k_2-1)}= \dots=\\
    &\frac{1 - p(a_j|\mathcal{H}_{s-1})}{p(a_j|\mathcal{H}_{s-1})(k_1-1)}= \dots = \\
    &\frac{0 - p(a_N|\mathcal{H}_{s-1})}{p(a_N|\mathcal{H}_{s-1})(k_2-1)}
\end{split}
\end{equation}
Our proof is completed by showing (\ref{eq:otherJ}) and (\ref{eq:correct}) are correct.
\begin{equation}
\label{eq:otherJ}
    \frac{- p(a_i|\mathcal{H}_{s-1})}{p(a_i|\mathcal{H}_{s-1})(k_2-1)}= \frac{1 - p(a_j|\mathcal{H}_{s-1})}{p(a_j|\mathcal{H}_{s-1})(k_1-1)}, \ \ \forall i\in [1,N], \ i\neq j
\end{equation}
\begin{equation}
\label{eq:correct}
    \frac{- p(a_i|\mathcal{H}_{s-1})}{p(a_i|\mathcal{H}_{s-1})(k_2-1)}= \frac{- p(a_q|\mathcal{H}_{s-1})}{p(a_q|\mathcal{H}_{s-1})(k_2-1)} \ \ \forall i, q \in [1,N], \ i,q\neq j
\end{equation}
Trivially, (\ref{eq:correct}) is correct. Using (\ref{eq:k1k2}), (\ref{eq:otherJ}) is also correct.
\begin{equation}
    \frac{- 1}{k_2-1}= \frac{1 - p(a_j|\mathcal{H}_{s-1})}{p(a_j|\mathcal{H}_{s-1})(k_1-1)} \\[.2in]
\end{equation}
\begin{equation}
\begin{array}{rl}
    p(a_j|\mathcal{H}_{s-1})(1-k_1)&= (1-p(a_j|\mathcal{H}_{s-1}))(k_2-1) \\[.2in]
    &=  (1 - p(a_j|\mathcal{H}_{s-1}))\left(\displaystyle{\frac{1-p(a_j|\mathcal{H}_{s-1})k_1}{1-p(a_j|\mathcal{H}_{s-1})}} - 1 \right)\\[.2in] 
    &=  (1 - p(a_j|\mathcal{H}_{s-1}))\\
    & \ \left(\displaystyle{\frac{p(a_j|\mathcal{H}_{s-1})-p(a_j|\mathcal{H}_{s-1})k_1}{1-p(a_j|\mathcal{H}_{s-1})}}\right) \\[.2in]
    & = p(a_j|\mathcal{H}_{s-1})(1-k_1)
    \end{array}
\end{equation}
\end{proof}
\subsection{Proof of Lemma 1}
\textbf{Lemma 1.}
\textit{
For $\tau \geq \displaystyle{\frac{1}{|\mathcal{A}|}}$ and $|\mathcal{A}| > 1$, decision boundaries $p(a)=\tau$ and $H(X) \leq \tau''$ intersect when entropy is at its maximum $\tau''$, at the midpoint of line $p(a)=\tau$, $P_m$, where $p_{m_k}, \forall k \in \mathcal{A}$ is defined as follows;
\begin{equation*}
p_{m_k} =
  \begin{cases}
    \tau       & k = a\\
    \displaystyle{\frac{1-\tau}{|\mathcal{A}|-1}}  & k \neq a
  \end{cases}
\end{equation*}
and 
\begin{equation*}
\label{eq:intersection}
    \begin{array}{rl}
        \tau'' &= -\tau\log{\left(\tau\right)}-(1-\tau) \log{\left(\displaystyle{\frac{1-\tau}{|\mathcal{A}|-1}}\right)}. \\[.1in]
        \end{array}
\end{equation*}
}
\begin{proof}
The entropy is upper bounded by a strictly concave function of the maximum likely element of set i.e., $p(x^*)$ and cardinality of the set, where the function is monotonically increasing for $\forall x\in[1/|\mathcal{A}|,1]$ where $|\mathcal{A}| \geq 1$.
\begin{equation}
\label{eq:entropy_bound_sup}
\begin{split}
    H(X) &= -\sum_x p(x)\log(p(x)) \\
    &= -p(x^*)\log(p(x^*)) - \sum_{x\neq x^*} p(x)\log(p(x)) \\
    &\leq -p(x^*)\log(p(x^*))-(1-p(x^*))\log{\left(\displaystyle{\frac{1-p(x^*)}{|\mathcal{A}|-1}}\right)} \\
\end{split}
\end{equation}
Considering $x^*=a$, $p(x^*) = p(a) = \tau$, according to \eqref{eq:entropy_bound_sup}, we have
\begin{equation}
\label{eq:entropyValue}
    \begin{array}{rl}
        \tau'' &=-\tau\log{\left(\tau\right)}-(1-\tau) \log{\left(\displaystyle{\frac{1-\tau}{|\mathcal{A}|-1}}\right)} \\[.1in]
        
        \end{array}
\end{equation}
and
\begin{equation}
\label{eq:entropyValue}
    \begin{array}{rl}
        P_m &= \displaystyle{\arg \max} \{ H(X) \} \\[.1in]
        p_{m_k} &=
  \begin{cases}
    \tau       & k=a\\
    \displaystyle{\frac{1-\tau}{|\mathcal{A}|-1}}  & k\neq a
  \end{cases}
        
        \end{array}
\end{equation}
where $P_m$ is the midpoint of line $p(a)=\tau$ in the simplex.
\end{proof}
\end{document}